\documentclass[preprint,12pt,3p]{elsarticle}

\usepackage{amssymb}

\usepackage{graphicx}

\usepackage{caption}

\usepackage{subcaption}

\usepackage{hyperref}

\usepackage{color}

\usepackage{multirow}

\usepackage[normalem]{ulem}

\useunder{\uline}{\ul}{}

\usepackage[ruled]{algorithm2e}

\usepackage{rotating, graphicx}

\usepackage{longtable}
\usepackage{lscape}
\usepackage{pdflscape}

\usepackage{amssymb}
\usepackage{amsmath}

\usepackage{hyperref}

\usepackage{multirow}

\usepackage{caption}

\usepackage{subcaption}

\usepackage{color}

\usepackage[table,xcdraw]{xcolor}

\usepackage{rotating, graphicx}

\usepackage{xcolor}

\usepackage{adjustbox}

\journal{arXiv}

\begin{document}

\begin{frontmatter}

\title{On the Image-Based Detection of Tomato and Corn leaves Diseases : An in-depth comparative experiments}

\author[label1]{Affan Yasin}
\ead{affan.yasin@tsinghua.edu.cn}

\author[label2]{Rubia Fatima \corref{cor1}}
\ead{rubiafatima91@hotmail.com}




\address[label1]{School of Software, Northwestern Polytechnical University, Xian, 710072, Shaanxi, China}
\address[label2]{School of Software, Tsinghua University, Beijing, P.R.China}

\cortext[cor1]{Corresponding Author : Rubia Fatima} 






\begin{abstract}
The Convolutional Neural Networks (CNN) have done remarkable achievements in the area of image categorization. The research article aims to provide a novel plant disease detection model based on plant image classification. The new training approach and methodology used to allow a simple and fast system implementation in practice. The created model is capable of classifying two distinct kinds of plant diseases into four distinct classes. To our knowledge, this technique for identifying plant diseases is being presented for the first time. In the \textbf{\textit{experiment number 1}}, we have used the four models namely Inception\mbox{-}V3, Dense\mbox{-}Net\mbox{-}121, ResNet\mbox{-}101\mbox{-}V2 ,and Xception model to perform the CNN training. The new plant disease image dataset we have created contains 1963 total images of tomato plants taken from the PlantVillage dataset. We have used 1374 for training and 589 for testing/validation of the images. Also, we have created containing 7316 total images of corn plants taken from the \textit{PlantVillage} dataset, and we have used 5121 for training and 2195 for testing/validation of the images. The first experiment results show that Xception model has better val\_accuracy and val\_loss as compared to the other three models for the tomato and corn dataset. The evaluated values for the tomato and corn dataset are ``val\_accuracy = 95.08\%, and val\_los = 0.3108", and ``val\_accuracy = 92.21\%, and val\_los = 0.4204" respectively. In \textbf{\textit{experiment number 2}}, we have utilized CNN using Batch Normalization for disease detection in Tomatoes and Corn leaves. The results for detection of diseases for both the vegetable leaves are around 99.89\% (training set) and for val\_accuracy\textgreater=97.52\% with val\_loss= 0.103. In \textbf{\textit{experiment number 3}}, CNN architecture was used to classify as the base model. As a next step, more layers were added in the baseline model (Model 2). Furthermore, skip connections were added in model 3, and in model 4, regularizations were added. The detail of the experiment and models efficicency is shown in the paper (sub-section 1.5). In 
\textbf{\textit{experiment number 4}}, all the images of corn and tomato's were combined and further run different model. Dataset of 17603 were used for training, and 4409 images for testing purposes. Furthermore, different models such as, MobileNet (val\_accuracy= 86.73\%), EfficientNetB0 (val\_accuracy= 93.973\%), Xception (val\_accuracy= 74.91\%), InceptionResNetV2 (val\_accuracy= 31.03\%), and CNN (59.79\%)  were used. Besides that, we have also proposed our model with val\_accuracy= 84.42\%.

\end{abstract}


\end{frontmatter}






\section*{Introduction}

With an ever-increasing population around the globe, the global food system has been subjected to certain threats. According to an estimate, the global population will reach 9 billion by 2050 which can be an alarming scenario for the global food supply. This is the reason the efforts to improve agriculture sustainability have been given wide attention. The sustainable development goals of the United Nations put special focus on sustainable agriculture as it is highly crucial for global food and nutrient requirements \cite{1jones2016new}. Moreover, for many countries worldwide, agriculture is also an important economic sector and a source of employment for millions of people. The countries that are primarily dependent upon agriculture, the smooth running of everything in the country is also dependent upon this sector \cite{2panigrahi2020maize}. Any unexpected disruption in agriculture sector due to disease attack have grave implications for developing economies. Tomato is one the most widely used vegetable crop which is used for cooking purposes as well as raw food material \cite{3giovannoni2007fruit}. Tomato is also considered a highly nutritional vegetable crop. It contains antioxidants, vitamins, dietary fibers as well essential minerals. From a more scientific perspective, it has also been utilized as an ideal crop for the evaluation of genetic model studies aimed at improving crop production \cite{4chen2011glycinebetaine}. Tomatoes are a major vegetable crop globally and represent about 15\% of the global vegetable consumption at an annual per capita of around 20 kg. Fresh tomato production worldwide reaches 170 million tons per year, making it one of the most produced vegetable crop \cite{harvey2002exploring}. Tomatoes are mostly manufactured in the USA, India, Turkey, Egypt and China. Tomato disease, according to survey figures of the Agriculture Organization and United Nations Food, is the main cause of a global tomato crop decrease with an annual loss rate of 8\% to 10\% \cite{1heuvelink2018tomatoes,doi:10.1080/08839514.2017.1315516}. However, the majority of tomato illnesses begin on the leaves and subsequently spread throughout the plant. Automatically identifying tomato plant diseases correctly may assist in improving tomato production management and offers a favorable growing environment \cite{2nabeel2019cas,ashqar2018image}.

Corn is a high annual grass with a strong, erect and robust stem. The large, wave-margined leaves are separated from one other on opposite sides of the stem. On the tassel ending the main axis of the plant, stamina (male) blooms are formed. Pistillate (female) inflorescences, edible ears have a lengthened axis and longitudinal ranges of the pairs of spikelet's, producing two ranks of grain, each with pairs of spikelet's. The most common form of yellow and white maize is the type of food, although red, blue, pink and black kernels are often banded, speckled and striped \cite{3ranum2014global}. Each ear has modified leaves known as husks or shucks \cite{4zhoudr}.

Plant crop has always been prone to disease attacks across various countries. The highest frequency of diseases is because of disease-causing agents such as fungi, bacteria and virus \cite{5kartikeyan2021review}. Moreover, various environmental factors (humidity, temperature, and rain) play a crucial part in the outbreak of a particular tomato disease. Climate change is another factor that has been responsible for the increasing magnitude of disease attacks \cite{6verma2018prediction} due to the disturbances in rainfall and temperature patterns. Studies have shown that it can augment the rate of pathogen development and can also reduce host resistance against the pathogen \cite{7coakley1999climate}. Diseases serve as a major limiting factor to its potential yield and growth which consequently affects the agricultural economy of a country \cite{8hanssen2012major}. For instance, in the past years, the tomato crop has been negatively impacted by an early blight which has been considered as one of the most common vegetable plant diseases. This disease caused a huge amount of yield and production losses in many other fruits along with the tomato crop. In the same manner, late blight has been a significant cause of considerable yield losses and its negative impact has been even worse in changing climate conditions, such as the unexpected increase in humidity \cite{9brahimi2017deep}. It is a quite daunting situation for farmers who invest a lot of time, money and energy is raising the crop with expectation for optimal output.

The expense and subjective risk of misjudgment are traditional expert diagnostics for tomato plant disease. Computer vision deep learning and, machine learning in agricultural bug diagnosis is quite common with the fast growth of computer technology \cite{5latif2020integration,loey2020deep}. Color, textures or form characteristics segment traditional machine vision techniques into RGB images of crop disease. But the features of various illnesses are the same; therefore, the kinds of illnesses are difficult to measure. In a challenging natural setting, the accuracy of identification of sickness is low. Therefore, it seems quite pertinent to make efforts to detect and identify plant diseases through early monitoring so that proper treatment could be applied before things get worse in terms of disease spread. Accurate disease detection and identification at an early stage is also a major contributing factor to the accomplishment of sustainable agriculture. Identification and detection of tomatoes at an early phase of growth is necessary to protect the quality and quantity of the crop  The most common method generally used for the detection and identification of the disease is through observations. Experts employ manual methods and approaches to match the symptoms with a particular pathogen attack and then determine the disease \cite{6verma2018prediction}. However, manual detection of the plant disease is a time-consuming as well a hectic task that would require continuous monitoring by knowledgeable personnel. In the case of large farms, this kind of disease identification would need a large number of employees and would entail a high cost \cite{5kartikeyan2021review}. Traditional methods for detection also require skilled technicians and a complete laboratory setup.  Last but not least, the observation of farmers may be subjected to error and misjudgment as there will always be the need for experts to incessantly monitor the plant symptoms \cite{10al2011fast}. This scenario iterates the need for an alternative that is more quick, cost-effective and accurate.


Convolutional Neural Networks (CNN) is now extensively utilized in crop disease detection for actual agricultural settings with CNN. Deep Learning (DL) technique is a relatively new technique in Machine Learning and is largely used on large and complex amounts of data also known as Big Data. It has been used by scientists and researchers to exploit raw and unstructured data directly without using manual and traditional methods \cite{9brahimi2017deep}. In Deep Learning, ANNs and DNNs have been used to develop various powerful architectures, such as Inception V3, ResNet 50, ResNet 18, etc., which have been employed for image classifications in agriculture for disease detection of various plants \cite{11he2016deep,12szegedy2016rethinking}.

Many pieces of research using profound training technology have been conducted to increase the overall survival of vegetables, fruit and fields by early identification of diseases and the subsequent handling of diseases. Zhang et al \cite{8hang2019classification}, applies the transfers of the original Alex network, and the ten types of tomato leaves are better recognized on average \cite{8hang2019classification}. Rangarajan et al. \cite{9rangarajan2018tomato} approximately 97\% of all seven separate tomato leaves utilize the original AlexNet, VGG 16 network topology, and Migration learning. The impact on the accuracy and speed of illness diagnosis is evaluated by weight, deviation and rate learning \cite{9rangarajan2018tomato}.  Fuentes et al. cameras of various resolutions utilize Faster RCNN, R-FCN and SSD for training purposes to collect the images of 9 tomato illnesses and insect pests \cite{10fuentes2017robust}. Long et al. has taught AlexNet and Google Net using Camellia Oleifera illnesses detection \cite{11long2018image}. Karthik et al. utilize a ResNets pre-trained network to categories 98.8\% of 7 tomato illnesses \cite{12karthik2020attention}. Safonova et al. presented a profound model detection structure for tomato plant illnesses, the residual networks refined and upgraded and transfer learned to acquire key diseases. Although transferring education may lead to better recognizing outcomes, the original AlexNet and VGG16 networks do not satisfy the current application and deployment of the model with complicated structures and several parameters \cite{13safonova2019detection}.

Although the profound theory of learning has produced excellent results in categorizing images, it also has excessive progression and over-fitting issues. In this light, this article brings into the architecture of the deep learning network a spacious representation of the excellent linear decomposition ability and the strong structural benefits of multilayer, nonlinear cartography. This paper is used to make full use of the sparse representation \cite{14pouyanfar2018survey}. It completes the approximation of complicated functions and creates a deep model of learning using adaptive approaches. The deep learning model resolves the issue of the function approach. At the same time, the optimized kernel function is suggested as a sparse representation classification technique to replace the deep learning model classifier \cite{15shrestha2019review}. The categorization impact will be improved. This article categorizes images based on the non-negative, diminished representation of the stacked sparse coding depth learned by optimizing the model function kernel \cite{16imran2020smart}. The originality of this article is to build a deep learning model with the capacity to adapt. This article simultaneously offers a novel sparse classification system for optimizing kernel functions to replace the classifier in the profound learning model \cite{17bejnordi2017diagnostic}. The sick region only fills a portion of the image size for the diagnosis of tomato and maize plants. This research adds an attention module to the basic CNN network model to get information from a complicated environment on key illness characteristics automated. The function extraction focuses on the illness channel and eliminates incorrect information on the function channel. A new CNN network model to correctly diagnose various plant diseases from tomatoes has been suggested in this document.

Therefore, the current paper aims to present deep learning as a feasible approach for classifying the diseases in tomato plants based on the images of the disease affected leaves. The study develops a deep model by using the Deep Neural Networks approach. The study aims to make the use of the developed model in other tasks where the data-set is even larger and more complex. Moreover, it aims to enable the deep model to efficiently use raw data-sets and outperform previous models utilized for disease detection and classification. This paper also focuses on underlining the specific functions of all ANNs and their use in a data-set that mainly comprises a number of disease images of tomato and corn plants. Moreover, this study intends to highlight the best architecture for disease classification which can be an invaluable help turn CNN models into an automated system for disease classification and detection, helping farmers and experts to tackle tomato and corn diseases at an early stage. 

\vspace{3mm}

This paper summarizes the major contributions as follows:

\begin{enumerate}
    \item For different tomato plant diseases in healthy plants to satisfy the diagnostic criteria. The capacity to generalize and adjust the model in real applications is also enhanced via data augmentation techniques.
    
    \item This study contains the multi-scale topology network of CNN for the detection of tomato-plant illness. A multi-scale extracting module is added, depending on the remaining block.
    
    \item This study created a multidisciplinary dependence connection between the three-dimensional (C, H, W) characteristic map of the retrieved tomato and corn-plant disease and geographical data with minimal calculation. It may provide useful lesion characteristics in a complicated environment and distinguish contextual information.

\end{enumerate}

\begin{figure*}[!ht]
\begin{center}
\includegraphics[width=0.70\textwidth]{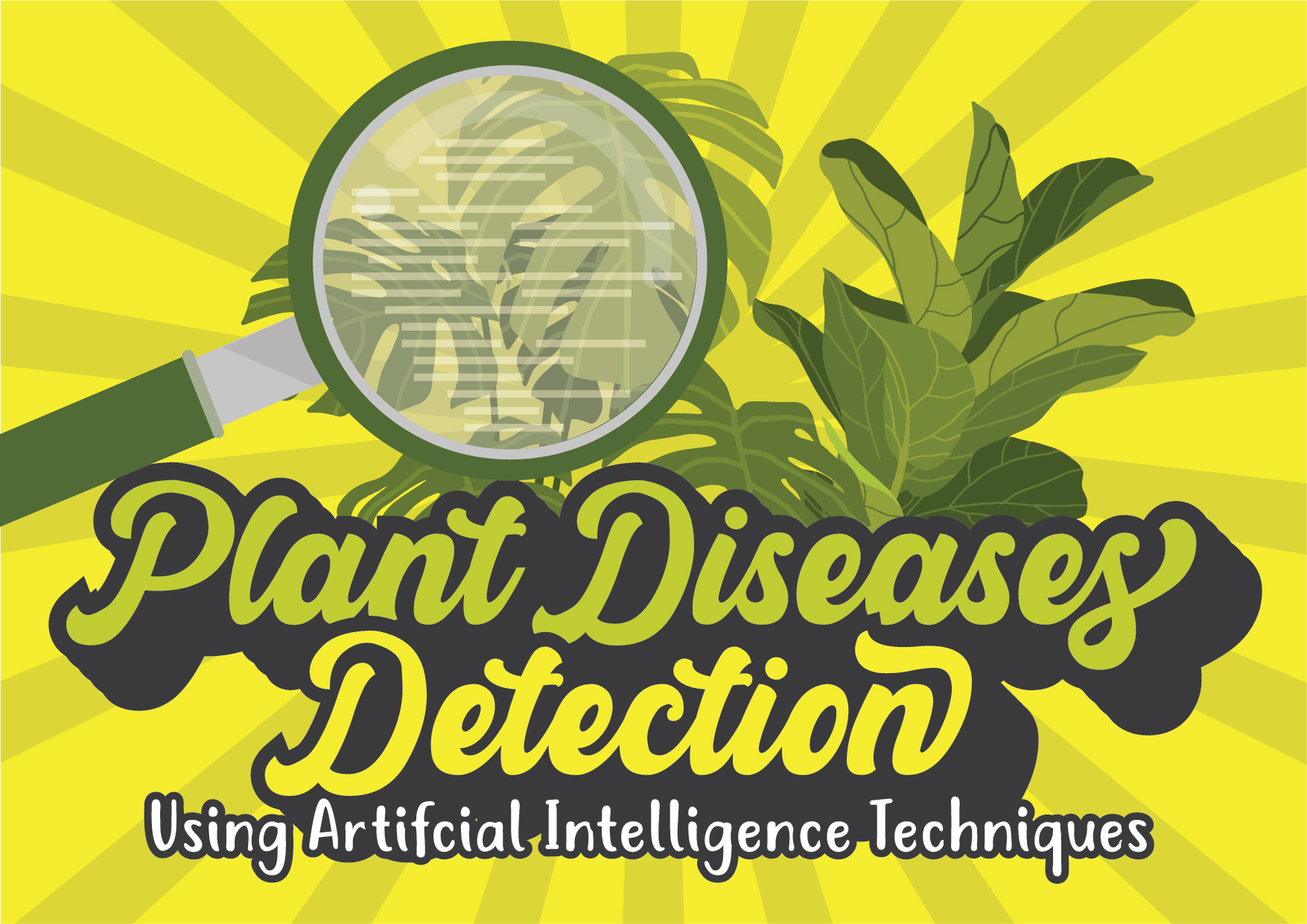}
\caption{Plant Diseases Identification Using Artificial Intelligence Techniques }
\label{figs:pdiuait}
\end{center}
\end{figure*}

In our research study, we have five sections. The first section explains the background and introduction to our research topic. Furthermore, the second section describes the literature related to our research study (literature review). The third section discusses the material and methods used in the study. The fourth section debates the Experimental Evaluations, and finally, section 5 concludes the paper.

\section*{Literature Review}

Deep learning has greatly advanced the area of machine learning in general and particularly in computer vision. Deep learning has been applied in numerous domains, resulting in substantial machine learning and computer vision achievements. The use of deep education in agriculture in the field of computer usage was just recently done. Profound knowledge in identifying herbal diseases enabled greater prediction accuracy and increased recognized diseases and plant species \cite{25loey2020deep}. Recently, extensive study has focused on developing a fast, automatic and accurate picture and categorization system. This study aims at refining, evaluating and classifying the deep neural grids, state-of-the-art, for pictorial plant diseases. Dense-Nets continually improve their accuracy with an increasing number of repetitions, without over-fitting or performance deterioration \cite{26too2019comparative}.

A fresh dataset with 79,265 images has been added by Arsenovic et al. to make the data collection with plant images the biggest. In varied weather, from different perspectives, and during daylight hours, images were shot with an inconsistent history imitating a real scenario. A new two-stage design for a neural network was suggested to categorize plant disease centered on a true environment. The trained model reached 93.67\% accuracy and may be utilized to detect illnesses in a controlled setting \cite{28arsenovic2019solving}.

Atila et al. \cite{18atila2021plant} suggested the Efficient-Net deep learning design in the categorization of plant disease. Compared to other state-of-the-art deep learning models, the performance of this model. Test results indicate that Efficient-Net architecture models of the B5 and B3 achieved the highest values with 99.91\% and 99.97\%, respectively, compared to other CNN models.

Farmers cannot just detect the illness by looking at the crop plant since the healthy crop and the crop in question seem to be the same in the early stages. Deep learning models can address this problem. There are 20 data gathered from the dataset of PlantVillage for the research study. The technique presented provides 99.6\% maximum fine-tuning accuracy of ResNet50 \cite{19sravan2021deep}.

Delnevo et al. \cite{20delnevo2021deep} envisages the Social Internet of Things to detect and communicate environmental variables (sunlight, humidity, air temperature, soil water). The authors suggest deep learning for the identification of plant diseases using crowd-sourcing to gather, classify images, the architecture, the profound concept of learning and the responsive web app. Some tests and usability tests, final comments, constraints, and future work are presented and discussed.

The use of Convolutional Neural networks (CNNs) for disease classification in plants has been increasing since its emergence. Researchers have been indulged in an effort to develop a reliable system framework through which a large number of classes can be developed, which still lagging in need of a substantial amount of study. Researchers have discovered a novel method of detecting illness using computer vision and profound learning in leafy plants. The scientists discovered that SVM is one of the good methods for detecting plant diseases. The authors believe they can help prevent illness elsewhere, including Africa, by using their approach \cite{23mohameth2020plant}.


The purpose of this study is to identify illnesses in tomatoes or greenhouses of plants. Deep learning to identify different illnesses on the tomato plant leaves has been utilized for this purpose. The objective of the research was to execute the deep learning algorithm on the robot in real-time. Thus, when the robot goes manually or autonomously on or in the greenhouse, it can identify diseases \cite{32durmucs2017disease}.

Sladojevic et al. \cite{13sladojevic2016deep} conducted a study in which they used Deep Neural Networks (DNNs) to develop the latest approach for the accurate recognition of disease which would be based on the classifications of leaf images. Authors developed model helped to detect 13 different diseases in the selected plants. In order to execute CNN training, they employed the Caffe framework. Their experimental model revealed an accuracy of about 96.2\%.
Prasanna et al. \cite{14prasanna2016using} collected a dataset of 54306 images of both infected and healthy plant leaves to train a deep CNN for identifying the plant diseases as well as species. For the total 26 diseases present in the images, their model depicted a precision of about 99.35\%. They concluded that such an approach is highly helpful in promoting smart-phone based detection of plant diseases.

Ferentinos et al. \cite{15ferentinos2018deep} developed CNN models with an aim to detect and diagnose disease in plants by using leaves of disease infected plants and healthy plants. The author employed deep learning technique and the models were trained with the help of 87,848  images having 25 different plants in 58 discrete classes. In the study, five different training CNN architecture models were used. The best performing model for disease detection reached an accuracy of about 99.52\%.

Fuentes et al. \cite{16fuentes2017robust} aimed to find an appropriate CNN architecture among the used deep learning meta architectures for the detection of real-time diseases in tomato plants. To this end, they used deep learning-based approach while using the plant images with varying resolutions. They employed three architectures and combined them with deep feature extractors, e.g. ResNet. After training and testing their model on the complex images of tomato diseases, they found that their model has the ability to identify nice diseases from a challenging dataset of plant disease images.

Wang et al. \cite{17wang2017automatic} developed a model to detect the severity of the disease. To this end, they used various deep CNNs and these networks were tested and trained by transfer learning. The results showed that the best deep model for the disease severity diagnosis is the VGG16 deep model which showed an accuracy of 90.4\% showing its potential for disease detection for sustainable agriculture.

Lu et al. \cite{18lu2017identification} conducted a study to propose a model of disease identification in rice based on deep convolutional neural networks by using high resolution images dataset of rice plants leaves and stems. They employed a ten-fold cross-validation strategy for the experiment. The results exhibited that the model based on deep CNNs showed an accuracy of about 95.48\%. They concluded that the accuracy of CNN-based models is quite higher than the traditional models and methods of detection. Moreover, simulation output further revealed the effectiveness of the model developed in the research study. 

Amara et al. \cite{22amara2017deep} also implied the use of a deep learning approach to classify image datasets of banana disease in an automated manner. They found that their model was able to classify and detect plant diseases even if the image resolutions are contrasting or the background is complex, depicting the accuracy of its operation.

In empirical research, Too et al. \cite{23too2019comparative} employed the use of deep convolutional neural networks (CNNs) in order to detect plant diseases based on images of varying resolutions of the diseased leaves. They aimed to compare various deep learning architectures to pinpoint the best model for disease detection. They evaluated four distinct deep architectures among which DensNet depicted the highest tendency of increased accuracy in a consistent manner. They used Keras in combination with Theano to train the architectures under study. Moreover, its testing accuracy outperformed all other four architecture with 99.75\% accuracy, depicting its appropriateness for disease detection and classification. Saleem et al. \cite{24saleem2019plant} reviewed the implementation of deep learning based models for the identification and detection of plant diseases. They argued that some gaps still exist in the image based use of deep learning for the disease detection for greater transparency.

Begum et al. \cite{25begum2020diagnosis} designed a model based on deep convolutional neural networks (CNNs) for the identification and detection of 15 classifications of diseases. The dataset consisted of images of three different crops, viz tomato, potato and pepper bell with varying resolutions. For this purpose, they employed distinct deep learning architectures. The testing and training of the architectures revealed that Modified Mobilenet was the best deep CNN model for the detection and classification of diseases, depicting high accuracy. Overall efficiency was recorded at around 97 percent.

Saleem et al. \cite{26saleem2020image} conducted a study to detect different types of diseases by using three distinct deep learning meta-architectures. The leaf images of healthy and diseased plants served as the training dataset. Learning optimizers were also employed to enhance the precision of the final detection results. The final results showed that all three models had high detection accuracy. However, SSD trained model had the highest disease detection accuracy whivh was about 73.01\%.

In a recent study, Rao et al. \cite{27rao2021plant} developed a novel approach using deep convolution networks on the basis of the classification of the diseased leaf images of varying resolutions. They aimed to accurately identify and detect the plant disease to protect the crop at an early stage. They used the language of Python for deep CNN. They found that their developed model had high accuracy for detecting about 13 distinct diseases and differentiate between diseased and healthy plant leaves. In exploratory research, Barbedo et al. \cite{28barbedo2018factors} focused on highlighting the factors that influence the efficiency of deep convolutional neural networks for the detection of plant diseases in image-based detection methods. The author found that many intrinsic, as well as extrinsic factors, impact the accuracy of deep CNNs, such as image background, conditions during image capturing, symptom segmentation, and symptom variations.

Kawasaki et al. \cite{29kawasaki2015basic} proposed a novel approach for plant disease detection by using deep convolutional neural networks (CNNs). For this purpose, they used 800 images of cucumber leaves with different resolutions for the training of the model and utilized a 4-fold cross-validation strategy. This assisted in expanding the data-set by developing extra images. They found that their model had an accuracy of about 95\% for the detection and classification of plants into diseased and healthy plants by using CNNs.

Lee et al. \cite{30lee2015deep} performed a study to develop a visualization model for the implementation of deep convolutional neural networks (CNNs) to detect and classify plant species based on high resolution images of diseased plant leaves. The model used the venation pattern of different plants for classification. They found that the developed model had high accuracy in determining and classifying the plant species.

Zhang et al. \cite{31zhang2018identification} utilized two improved deep learning-based models, i.e., GoogleLeNet and Cifar10, to identify and diagnose diseases in maize plants. The dataset comprised leaf images of nine different kinds of maize, including both healthy and diseased plants and then both models were trained. After the testing of trained models, they found that both improved models had high accuracy for disease detection and diagnosis, i.e., GoogleLeNet with 98.9\% average accuracy and Cifar10 with 98.8\% average accuracy. They concluded that such improved deep CNN models are quite invaluable in improving plant protection against diseases.

DeChant et al. \cite{33dechant2017automated} utilized the deep learning approach by employing deep convolutional neural networks (CNNs) to identify northern leaf blight (NLB) disease in maize. For this purpose, they used a limited image dataset of leaves containing NLB lesions to train the model. Then the testing was allowed to let the CNNs classify the disease images. They found that the deep learning-based model had high accuracy (96.7\%) on test set images for disease detection and classification of NLB disease. They concluded that such models can be used for the reduction of pesticide use as it will provide an understanding of the target amount of fertilizer needed for disease control. 

Fujita et al. \cite{34fujita2016basic} proposed a practical system for plant disease detection and classification by using deep convolutional neural networks (CNNs). They used 7520 images of health and diseased plant leaves of cucumbers as the dataset for the training of CNN architectures for the detection of viral diseases. They employed a 4-fold cross-validation strategy for the study. They found that their model depicted an accuracy of 82.3\% for disease detection and classification.

Ferentinos et al. \cite{35ferentinos2018deep} employed deep learning-based approach to detect and classify plant diseases in tomato leaves. Their proposed approach utilized two architectures as models, viz. inception V3 and convolutional neural network (CNN). The dataset consisted of images of diseased leaves with varying sizes and resolutions which was used for the training of the two models. The results revealed that CNN and inception V3 had accuracies of 94.72\% and 96.11\%, respectively. They concluded that such a model could be used for the early protection of tomato plants against diseases.

Islam et al. \cite{36islam2018jute} conducted a study to develop a technique for disease detection and classification using deep convolutional neural networks (CNNs). The dataset for the research entailed images of jute leaves including both healthy and diseased plants. The CNNs under study were trained using these images and were tested. They found that the technique had high accuracy for disease classification and detection and could be utilized to improve disease detection in other plants.

Huang et al. \cite{37huang2019development} conducted research to construct convolutional neural network models based on leaf images of disease infected and healthy plants for the classification and detection of plant diseases. The database consisted of 87848 images which were employed to train the model. The trained models were tested for detection. The results showed that model architecture had high disease detection accuracy (99.53\%). The author concluded that the developed model could be utilized to detect diseases at early stages and plant protection systems could be made more efficient.

\section*{Material and Methods}
This section thoroughly explains the entire deep CNN plant disease detection procedure. The process is broken down into several necessary steps, starting with gathering images for deep neural network classification.

\subsection*{Input Data}
It is necessary to use the proper datasets at every object classification stage, from the training stage up to the assessment of object recognition algorithms' performance. The whole collection of images from the PlantVillage\footnote{https://www.kaggle.com/abdallahalidev/plantvillage-dataset} dataset. The images in the dataset were divided into two categories; differentiate healthy leaves from sick leaves, a third category was added to the dataset. It is entirely composed of images of healthy leaves. A separate class containing backdrop images was added to the dataset to aid with classification accuracy. Thus, a deep neural network might be taught to discriminate between various types of leaves. The next phase was to add improved pictures to the dataset to round out the collection. The primary objective of the research, as the name suggests, is to form a network to discover the characteristics that differentiate one class from another. As a result, the likelihood of the network learning the right characteristics is enhanced by using more augmented images. Table~\ref{tabs:images} summarizes the number of original and enhanced images for each class that served as the disease classification model’s training and testing/validation datasets.

\begin{table}[h!]
\centering
\caption{Collection of images for the categorization of plant diseases}
\label{tabs:images}
\begin{tabular}{|l|c|c|c|}
\hline
\rowcolor[HTML]{FFFFC7} 
                & \textbf{\begin{tabular}[c]{@{}c@{}}Original \\ Images\end{tabular}} & \textbf{\begin{tabular}[c]{@{}c@{}}Training \\ Images\end{tabular}} & \textbf{\begin{tabular}[c]{@{}c@{}}Test / Validate\\ Samples\end{tabular}} \\ \hline
\textbf{Tomato} & 1963                                                                & 1374                                                                & 589                                                                        \\ \hline
\textbf{Corn}   & 7316                                                                & 5121                                                                & 2195                                                                       \\ \hline
\end{tabular}
\end{table}

\subsection*{Data Pre-processing Steps}

The images from the PlantVillage collection were collected in several formats, quality levels, and different resolutions. The final images for use as datasets for the deep neural network classification have been pre-processed to provide uniformity to enhance feature extraction. The image preparation technique also involved cutting all images manually, forming a square around the leaves to highlight the area of focus.

\begin{figure*}[!ht]
\begin{center}
\includegraphics[width=0.77\textwidth]{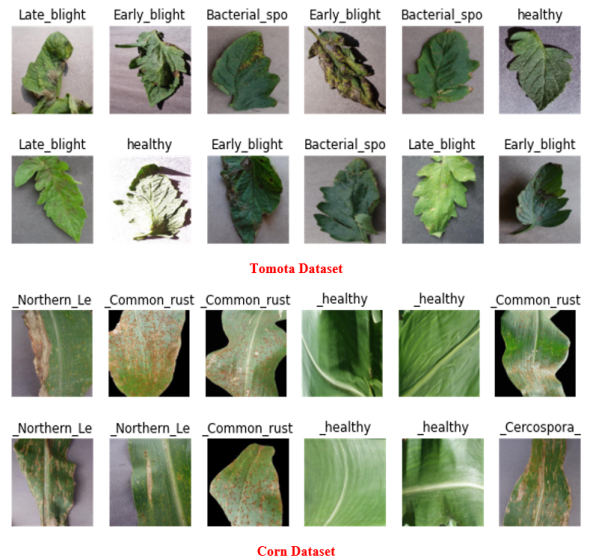}   
\caption{Sample screenshot of the labelled data set}
\label{figs:sampleshoot}
\end{center}
\end{figure*}

Images with a resolution of less than 500 px throughout the data collecting period were not considered genuine. Furthermore, only images with a higher area of interest resolution have been selected as eligible candidates for the dataset. As a result, the pictures were guaranteed to include all of the information necessary for feature learning. In order to minimize the training time, the pictures in the dataset were scaled automatically using the OpenCV framework, which was implemented in a Python script.

Numerous materials may be located via Internet searches; however, their relevancy is often suspect. Agricultural specialists reviewed plant images and labelled each image with the relevant illness term to verify the correctness of the classes in the dataset, which was originally categorized using keywords. As is well known, the training and validation datasets must include correctly categorized images. Only in this manner is it possible to build an adequate and reliable detection model, as shown in Figure~\ref{figs:sampleshoot}.

\subsubsection*{Data Augmentation Process}
This is a method simply to enhance the total number of labeled images in a detection study. There are many well-acknowledged methods of data augmentation, such as image rotation, vertical flip, horizontal flip, cutout, 90o counterclockwise flip, 180 degree rotation, 90 degree rotation, etc., \cite{krizhevsky2012imagenet}. In our study, we have utilized three unique data augmentation methods, including image rotation, brightness (increase or decrease) and contrast (enhancement or reduction). In Figure~\ref{figs:ima}, the examples or original image (Figure~\ref{figs:ima}-A), rotation (Figure~\ref{figs:ima}-B), brightness (Figure~\ref{figs:ima}-C) and contrast (Figure~\ref{figs:ima}-D) are illustrated.
The primary aim of increasing the dataset is to reduce overlap during the training stage by introducing a little distortion of the images. Overfitting occurs in machine learning and statistics when a statistical model reflects random noise or mistakes instead of an underlying connection. The image increase included many transformation methods, including refined transformation, the transformation of perspectives and basic image rotations. Affine transformations (linear transformations and vector add-ons) have been used to represent translation and rotation (e.g., all parallel lines in the original image remain parallel to those in the output image). Three points from the source image and their appropriate positions in the output image were required to find a transformation matrix. A transformation matrix was necessary for viewpoint transformation. Even after the change, straight lines would remain direct. Simple image rotating and rotation of the different axes by varying degrees were used to increase the image process.

\begin{figure*}[!ht]
\begin{center}
\includegraphics[width=0.60\textwidth]{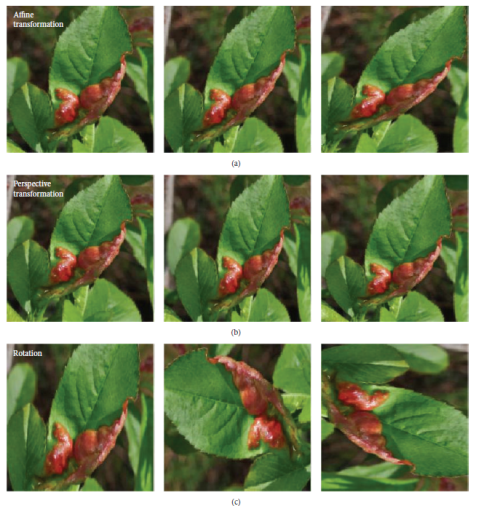}
\caption{Image modifications used for augmentation}
\label{figs:imagemodi}
\end{center}
\end{figure*}

\begin{figure*}[!ht]
\begin{center}
\includegraphics[width=0.75\textwidth]{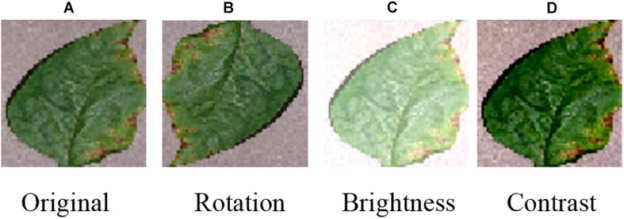}
\caption{Data Augmentation Method}
\label{figs:ima}
\end{center}
\end{figure*}

In the first row, the images are represented by the finished transformation in the single image; Images from different transformation viewpoints are displayed in the second row, in comparison to the input picture in the first row. In the final row, the basic rotation of an image. The increase method has been selected to satisfy the requirements; the plants may vary in visual perspectives in a natural setting, as shown in Figure~\ref{figs:imagemodi}. A custom program has been developed in C++ using the OpenCV library to automate the increasing process for many images in the collection and to modify transformation settings during runtime, which improves flexibility.

Regardless of the outcome, normalization of input is unnecessary; it is done after the first and second convolution layers due to its potential to lower error rates of top 1 and top 5. Neurons are split into “feature maps” inside a buried CNN layer. All neurons have the same weight and bias inside a feature map. In the function map, neurons look for the same functionality. These neurons are different because of their connections to other neurons in the lower layer.

The pooling layer, a type of nonlinear sampling, is another key layer of CNNs. Pooling provides translation invariance, works independently on each depth part of the input and spatially resizes it. Pooling overlapping is used to decrease overfitting. A drop-out layer is also employed in two completely linked layers to reduce overfitting. However, the failure to drop out means that training time is increased 2-3 times compared to a conventional neural network of accurate design. Bayesian Optimization studies have also shown the synergistic impact on ReLUs and drop-outs. It is implied that it is beneficial when combined.
\subsubsection{Fine-Tuning}
Fine-tuning is making tiny changes to improve or maximize results to enhance the effective or efficient process or function.


\subsection{Neural Network Models Architecture}

\begin{figure*}[!ht]
\begin{center}
\includegraphics[width=0.70\textwidth]{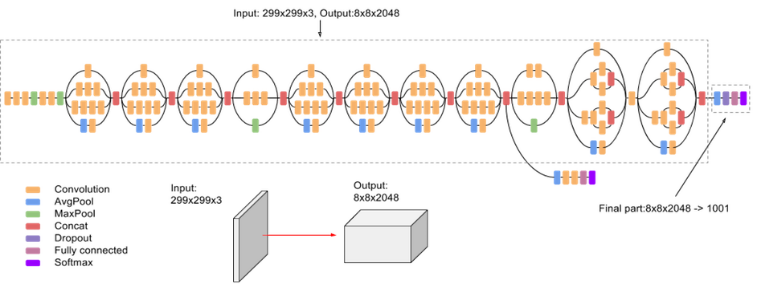}
\caption{A high-level diagram of the inception model \cite{34xia2017inception}}
\label{figs:inception}
\end{center}
\end{figure*}

\textbf{Inception v3} is an image recognition model commonly used which has been proven to achieve an exactness of over 78.1\% in the ImageNet dataset. The model culminates in numerous concepts that have been developed over the years by many scholars \cite{34xia2017inception}. The model consists of symmetry and asymmetry, including turbulence, average pooling, max pooling, concentrates, drop-outs and completely linked layers. Batchnorm is widely utilized and used for activation inputs throughout the model. Losses are calculated via SoftMax, as shown in Figure~\ref{figs:inception}.

\begin{figure*}[!ht]
\begin{center}
\includegraphics[width=0.70\textwidth]{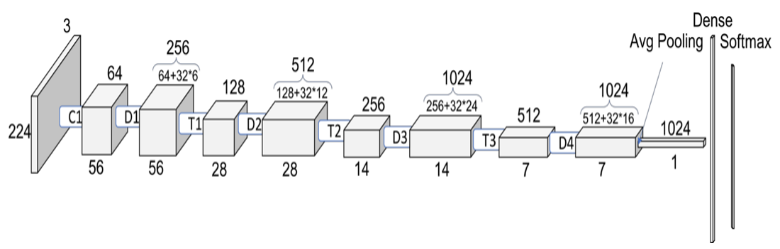}
\caption{A high-level diagram of the Dense-Nets model \cite{35o1991dense}}
\label{figs:Dense}
\end{center}
\end{figure*}

\textbf{Dense Net:} Convolutional networks are densely connected. This model has been chosen because Dense-Nets offer several strong advantages: they minimize the disappearance gradients, improve the propagation of features, promote reuse of functions and significantly lower parameters \cite{35o1991dense}, as shown in Figure~\ref{figs:Dense}.

\begin{figure*}[!ht]
\begin{center}
\includegraphics[width=0.70\textwidth]{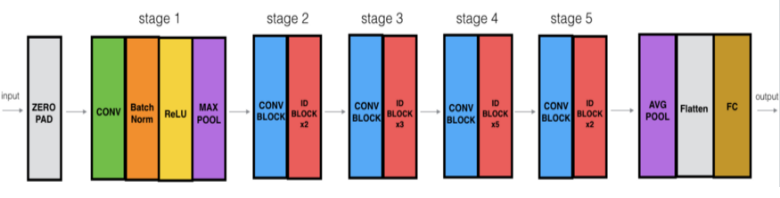}
\caption{A high-level diagram of the ResNet\mbox{-}101\mbox{-}V2 model \cite{36danilov2021real}}
\label{figs:v2}
\end{center}
\end{figure*}

\textbf{ResNet\-101\-V2} is a 101-layer deep convolution neural network. More than one million photos from the ImageNet dataset may be imported to train a pre-trained version of the network.. Images may be classified into 1000 different categories using the network's training data. It's because of this that the network now has a wide variety of capabilities. The network has a 224\mbox{-}by\mbox{-}224 pixel image input. Five stages make up the ResNet-101-V2 model, including blocks for identification and convolution \cite{36danilov2021real}. Each block has three levels of convolution, and each block of identity has three levels of convolution. Figure~\ref{figs:v2} shows the approximate number of trainable parameters for the ResNet 101-V2 around 23 million.

\textbf{Xception} is an expansion of the Architecture initiation that substitutes the Inception standard modules with wise, depth-specific Convolutions \cite{37chollet2017xception}, as shown in Figure 6.

\begin{figure*}[!ht]
\begin{center}
\includegraphics[width=0.70\textwidth]{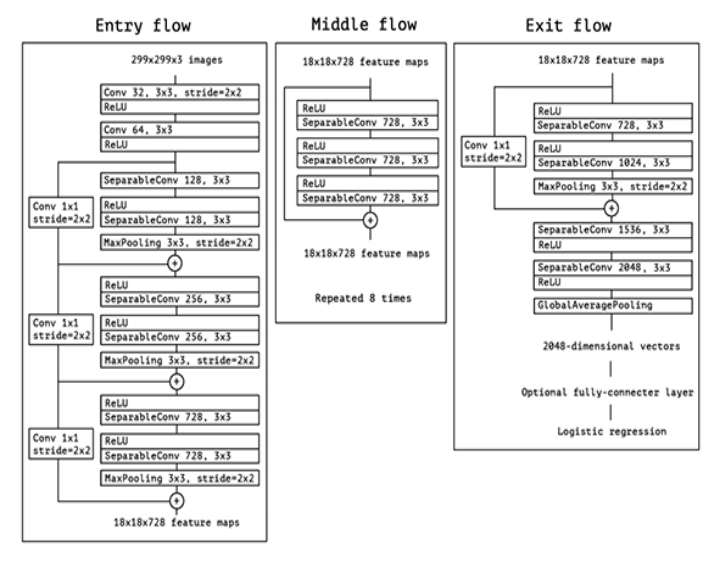}
\caption{ A high-level diagram of the Xception model \cite{37chollet2017xception}}
\label{figs:image6}
\end{center}
\end{figure*}

\section{Experimental Evaluations}

\subsection{Dataset Description}
\textbf{Data classification:} We have classified the data into multiple classes based on either Corn or Tomato. This classification is useful for the detection of their diseases. The dataset identified two crops which include:

\begin{itemize}
    \item Corn 
    \item Tomato
\end{itemize}

Both of these crops are classified according to the diseases and health conditions which are is as follows:

\textbf{Corn:} Corn dataset is classified into four distinct groups according to the following health condition and disease types:

\begin{itemize}
    \item Cercospora leaf spot 
    \item Common rust
    \item Healthy plants
    \item Northern leaf blight
\end{itemize}

The four classes of corn based on health conditions and diseases are shown in Table~\ref{tabs:class}. 

\textbf{Tomato:}
Similarly, the tomato dataset is classified into four distinct groups according to the following health condition and types of diseases: 
\begin{itemize}
    \item Bacterial spot
    \item Early blight
    \item Healthy plants
    \item Late blight
\end{itemize}

The four classes each of both tomato and corn are shown in Table~\ref{tabs:class}. 

\begin{table}[h!]
\centering
\caption{Classification of the disease Image dataset}
\label{tabs:class}
\begin{tabular}{|c|c|}
\hline
\rowcolor[HTML]{96FFFB}
\textbf{Classes}                                            & \textbf{Number of Classes} \\ \hline
Corn\_(maize)\_\_\_Cercospora ileaf ispot igray ileaf ispot & 1642                       \\ \hline
Corn\_(maize)\_\_\_Common irust                             & 1907                       \\ \hline
Corn\_(maize)\_\_\_Northern iLeaf iBlight                   & 1908                       \\ \hline
Corn\_(maize)\_\_\_healthy                                  & 1859                       \\ \hline
Tomato\_\_\_ iBacterial ispot                               & 570                        \\ \hline
Tomato\_\_\_ iEarly iblight                                 & 420                        \\ \hline
Tomato i\_\_\_Healthy                                       & 576                        \\ \hline
Tomato\_\_\_ iLate iblight                                  & 397                        \\ \hline
\end{tabular}
\end{table}

\subsubsection{Data Distribution/Shape:}
The shape of our data can be explained by the number of images each class has. 

\textbf{Corn:} Out of a total of 7316 images related to corn, 1642 images are illustrated by the class with Cercospora leaf spot and gray leaf spot disease. The other classes having common rust and Northern leaf blight diseases show 1907 and 1908 images, respectively. The healthy corn leaf images are found to be 1859.

\textbf{Tomato:} Out of a total of 1963 images in the dataset related to the tomato, 570 images are illustrated by class showing the bacterial spot disease. The other two classes having early blight and late blight diseases showed 420 images and 397 images, respectively. The class having healthy plants showed 576 images of the healthy tomato leaves.


\begin{figure*}[!ht]
\begin{center}
\includegraphics[width=0.70\textwidth]{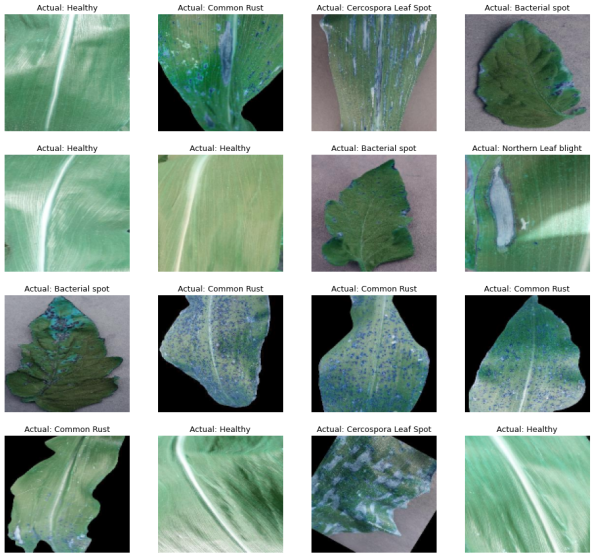}
\caption{Sample images of all classes of corn and tomato lead diseases}
\label{figs:sampleimage}
\end{center}
\end{figure*}

\subsection{Performance Evaluation}
We have reviewed the assessed outcomes from multiple deep learning methods, which we have used in the ``Tomato and Corn" dataset. The examined results demonstrate which model has the best accuracy and loss rate. The additional criteria we have used to validate the legitimacy of the deep learning model were precision, f1-score, and recall \cite{10.1007/978-3-540-31865-1_25} as calculated below:

\begin{equation} \label{eq1}
   Precision\  = \ True \ Positive/True \ Positive + False Positive
\end{equation}
\begin{equation} \label{eq2}
   Recall\   = \ True\ Positive / True\ Positive + False\ Negative
\end{equation}
\begin{equation} \label{eq3}
   F1 Score\  = \ 2*(Recall * Precision) / (Recall + Precision)
\end{equation}

\subsection{Experiment No 1 - Results - Using Inception-V3, Dense-Net-121, ResNet-101-V2, and Xception}

\subsubsection{The Evaluated Results for The Tomato Dataset}
In this section, we have used the dataset we have collected from the augmented images of the PlantVillage dataset. We have used the four classes of the tomato plant disease with the 1963 images plants. The classes we used for the analysis are ``Bacterial\_spot, Early\_blight, healthy, and Late\_blight". We have used different deep learning algorithms, ``Inception\mbox{-}V3, Dense\mbox{-}Net\mbox{-}121, ResNet\mbox{-}101\mbox{-}V2, and Xception", for the analysis.

\begin{table}[h!]
\centering
\caption{The Training and Testing Loss and Accuracy on Tomato Dataset}
\label{tabs:table2}
\scalebox{0.75}{
\begin{tabular}{|c|c|c|c|c|c|}
\hline
\rowcolor[HTML]{D9E2F3} 
\multicolumn{3}{|c|}{\textbf{InceptionV3}}                   & \multicolumn{3}{c|}{\textbf{Dense-Net-121}}                  \\ \hline
\rowcolor[HTML]{D9E2F3} 
\textbf{Epoch} & \textbf{Validation\_Loss} & \textbf{Validation\_Accuracy} & \textbf{Epoch} & \textbf{Validation\_Loss} & \textbf{Validation\_Accuracy} \\ \hline
1              & 29.3171            & 0.3430                 & 1              & 1.9240             & 0.6027                 \\ \hline
2              & 1.4428             & 0.6655                 & 2              & 4.3763             & 0.4907                 \\ \hline
3              & 1.3586             & 0.7963                 & 3              & 1.7133             & 0.7368                 \\ \hline
4              & 1.5484             & 0.8387                 & 4              & 0.7546             & 0.7759                 \\ \hline
5              & 2.2450             & 0.6740                 & 5              & 2.9624             & 0.5246                 \\ \hline
\rowcolor[HTML]{D9E2F3} 
\multicolumn{3}{|c|}{\textbf{ResNet-101-V2}}                 & \multicolumn{3}{c|}{\textbf{Xception}}                       \\ \hline
\rowcolor[HTML]{D9E2F3} 
\textbf{Epoch} & \textbf{Validation\_Loss} & \textbf{Validation\_Accuracy} & \textbf{Epoch} & \textbf{Validation\_Loss} & \textbf{Validation\_Accuracy} \\ \hline
1              & 86767.2943         & 0.2971                 & 1              & 4.8708             & 0.7216                 \\ \hline
2              & 25.2353            & 0.4363                 & 2              & 0.3334             & 0.9083                 \\ \hline
3              & 31.9316            & 0.2071                 & 3              & 0.4571             & 0.8727                 \\ \hline
4              & 13.0100            & 0.5382                 & 4              & 0.5380             & 0.9321                 \\ \hline
5              & 2.1292             & 0.6978                 & 5              & 0.3108             & 0.9508                 \\ \hline
\end{tabular}
}
\end{table}

\begin{figure*}[!ht]
\begin{center}
\includegraphics[width=0.95\textwidth]{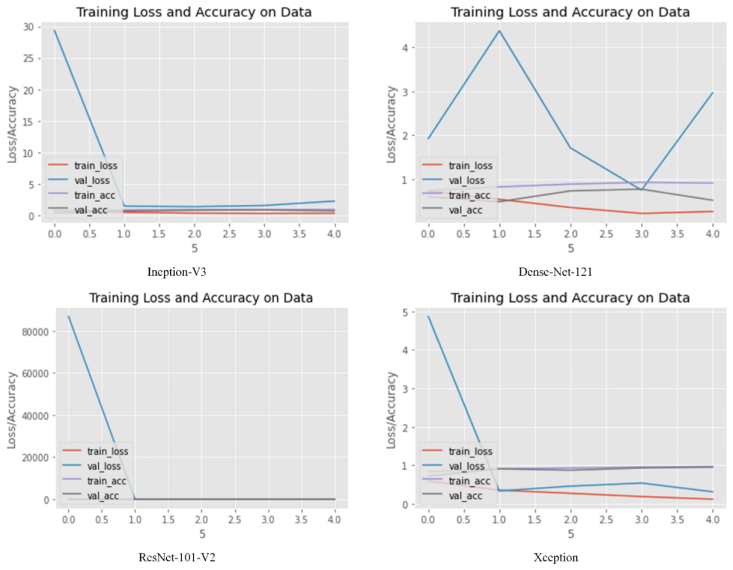}
\caption{The Graphical Representation for the Training Loss and Accuracy on the Tomato Data-set}
\label{figs:image7}
\end{center}
\end{figure*}

We have found that Xception's validation\_accuracy and validation\_loss are 95.08\% percent and 0.3108 percent better than any of the other tested models; the accuracy and loss rates are shown in Table~\ref{tabs:table2}. According to a table that lists the other models for this dataset, the Xception model outperforms all other models in terms of accuracy and loss rate. There is a graphical depiction of the accuracy and loss for all of our tomato dataset models in Figure ~\ref{figs:image7}.

The dataset we used is unbalanced, with 175 out of 589 occurrences (0.63, 0.83, 0.78, and 0.98) for various approaches falling within the bacterial spot category. The outcomes demonstrate that the predictor makes highly accurate predictions about any sample that is associated with the bacterial spot.



%





As shown in Table~\ref{tabs:table3}, we determined that the Xception model has high precision (95\%), recall, and F1-score as compared to other models.

\begin{table}[h!]
\centering
\caption{The Training and Testing Loss and Accuracy on Tomato Dataset}
\label{tabs:table3}
\scalebox{0.75}{
\begin{tabular}{|c|c|c|c|c|c|c|c|c|}
\hline
\rowcolor[HTML]{D9E2F3}
\multicolumn{1}{|l|}{\multirow{2}{*}{}} & \multicolumn{4}{c|}{\textbf{InceptionV3}}                                   & \multicolumn{4}{c|}{\textbf{Dense-Net-121}}                                 \\ \cline{2-9} 
\rowcolor[HTML]{D9E2F3}
\multicolumn{1}{|l|}{}                  & \textbf{Precision} & \textbf{Recall} & \textbf{F1-score} & \textbf{Support} & \textbf{Precision} & \textbf{Recall} & \textbf{F1-score} & \textbf{Support} \\ \hline
\textbf{Bacterial\_spot}                & 1.00               & 0.46            & 0.63              & 175              & 0.72               & 0.97            & 0.83              & 175              \\ \hline
\textbf{Early\_blight}                  & 0.45               & 0.92            & 0.61              & 129              & 0.67               & 0.05            & 0.09              & 129              \\ \hline
\textbf{healthy}                        & 0.92               & 0.28            & 0.43              & 121              & 0.34               & 0.91            & 0.50              & 121              \\ \hline
\textbf{Late\_blight}                   & 0.78               & 1.00            & 0.88              & 164              & 0.96               & 0.14            & 0.24              & 164              \\ \hline
\textbf{Accuracy}                       &                    &                 & 0.67              & 589              &                    &                 & 0.52              & 589              \\ \hline
\textbf{Macro avg}                      & 0.79               & 0.67            & 0.64              & 589              & 0.67               & 0.52            & 0.41              & 589              \\ \hline
\textbf{Weighted avg}                   & 0.80               & 0.67            & 0.65              & 589              & 0.70               & 0.52            & 0.44              & 589              \\ \hline
\rowcolor[HTML]{D9E2F3}
\multicolumn{1}{|l|}{\multirow{2}{*}{}} & \multicolumn{4}{c|}{\textbf{ResNet-101-V2}}                                 & \multicolumn{4}{c|}{\textbf{Xception}}                                      \\ \cline{2-9} 
\rowcolor[HTML]{D9E2F3}
\multicolumn{1}{|l|}{}                  & \textbf{Precision} & \textbf{Recall} & \textbf{F1-score} & \textbf{Support} & \textbf{Precision} & \textbf{Recall} & \textbf{F1-score} & \textbf{Support} \\ \hline
\textbf{Bacterial\_spot}                & 0.64               & 0.99            & 0.78              & 175              & 0.99               & 0.97            & 0.98              & 175              \\ \hline
\textbf{Early\_blight}                  & 0.66               & 0.18            & 0.28              & 129              & 0.92               & 0.92            & 0.92              & 129              \\ \hline
\textbf{healthy}                        & 0.83               & 0.43            & 0.57              & 121              & 0.89               & 0.90            & 0.90              & 121              \\ \hline
\textbf{Late\_blight}                   & 0.74               & 0.99            & 0.85              & 164              & 0.98               & 0.99            & 0.98              & 164              \\ \hline
\textbf{Accuracy}                       &                    &                 & 0.70              & 589              &                    &                 & 0.95              & 589              \\ \hline
\textbf{Macro avg}                      & 0.72               & 0.65            & 0.62              & 589              & 0.95               & 0.95            & 0.95              & 589              \\ \hline
\textbf{Weighted avg}                   & 0.71               & 0.70            & 0.64              & 589              & 0.95               & 0.95            & 0.95              & 589              \\ \hline
\end{tabular}
}
\end{table}

\subsubsection{The Evaluated Results for The Corn Dataset}

The dataset we gathered from the PlantVillage dataset's augmented photos was used in this part. With the 5121 plant photos, we used the four kinds of corn plant diseases. We used the classes `Cercospora\_leaf\_spot Gray\_leaf\_spot, Common\_rust, Northern\_Leaf\_Blight, and healthy" for the analysis. Different deep learning algorithms, including ``Inception-V3, Dense\mbox{-}Net\mbox{-}121, ResNet\mbox{-}101\mbox{-}V2, and Xception" have been utilized-forr the investigation. We compared them and determined which method would work best for real-time application.

\begin{figure*}[!ht]
\begin{center}
\includegraphics[width=0.99\textwidth]{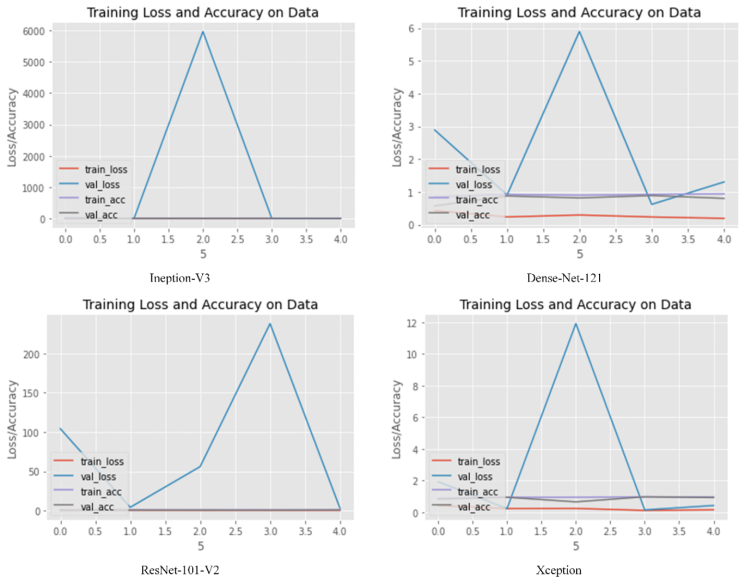}
\caption{The Graphical Representation for the Training Loss and Accuracy on the Corn Dataset}
\label{figs:image8}
\end{center}
\end{figure*}

The accuracy and loss rate of the various models we used for the analysis are shown in Table~\ref{tabs:table4}. From the results we have determined that the Xception model has the best val accuracy and val loss, which are respectively 92.21\% and 0.4204. As other models' accuracy and loss rates are included in the table, we have determined that the Xception model provides the comparatively better results for the corn dataset. The accuracy and loss for each model we constructed for the maize dataset are depicted graphically in Figure~\ref{figs:image8}.

\begin{table}[h!]
\centering
\caption{The Training and Testing Loss and Accuracy on Tomato Dataset}
\label{tabs:table4}
\scalebox{0.80}{
\begin{tabular}{|c|c|c|c|c|c|}
\hline
\rowcolor[HTML]{D9E2F3} 
\multicolumn{3}{|c|}{\textbf{InceptionV3}}                   & \multicolumn{3}{c|}{\textbf{Dense-Net-121}}                  \\ \hline
\rowcolor[HTML]{D9E2F3} 
\textbf{Epoch} & \textbf{Validation\_Loss} & \textbf{Validation\_Accuracy} & \textbf{Epoch} & \textbf{Validation\_Loss} & \textbf{Validation\_Accuracy} \\ \hline
1              & 16.5465            & 0.4665                 & 1              & 2.8874             & 0.5727                 \\ \hline
2              & 0.7263             & 0.8729                 & 2              & 0.9076             & 0.8765                 \\ \hline
3              & 5969.8509          & 0.2569                 & 3              & 5.8961             & 0.8150                 \\ \hline
4              & 0.3513             & 0.9034                 & 4              & 0.6201             & 0.8888                 \\ \hline
5              & 0.1415             & 0.9549                 & 5              & 1.3029             & 0.8018                 \\ \hline
\rowcolor[HTML]{D9E2F3} 
\multicolumn{3}{|c|}{\textbf{ResNet-101-V2}}                 & \multicolumn{3}{c|}{\textbf{Xception}}                       \\ \hline
\rowcolor[HTML]{D9E2F3} 
\textbf{Epoch} & \textbf{Validation\_Loss} & \textbf{Validation\_Accuracy} & \textbf{Epoch} & \textbf{Validation\_Loss} & \textbf{Validation\_Accuracy} \\ \hline
1              & 104.2504           & 0.1945                 & 1              & 1.9214             & 0.8196                 \\ \hline
2              & 4.1416             & 0.6437                 & 2              & 0.2276             & 0.9394                 \\ \hline
3              & 55.9583            & 0.5626                 & 3              & 11.9255            & 0.6519                 \\ \hline
4              & 238.0914           & 0.5367                 & 4              & 0.1412             & 0.9704                 \\ \hline
5              & 2.9520             & 0.7481                 & 5              & 0.4204             & 0.9221                 \\ \hline
\end{tabular}
}
\end{table}

The dataset we utilized is imbalanced, with 563 out of 2195 instances belonging to the class Common rust (0.97, 0.83, 0.88, and 0.93 percent for various techniques, respectively) and 563 out of 2195 examples belonging to the class Common rust As a result, your predictor nearly always correctly predicts any given sample corresponding to the class Common rust and achieves very high precision, recall, and f1-score for the class Common rust while achieving extremely poor precision, recall, and f1-score for the other classes. We have come to the conclusion that our model is incorrect when it comes to other classes, where your macro F1 properly reflects but weighted does not, thus resulting in the discrepancy between the two classes. We have evaluated that the Inception model has a high value of precision, recall, F1-score from other models, which are 95\% respectively, but the val\_loss of the model is high from the Xception model. So, we have derived that the Xception model has high precision, recall, F1-score from other models, 92\% respectively, as shown in Table~\ref{tabs:table5}.

\begin{table}[h!]
\centering
\caption{Average Evaluation performance of different models on training and testing data}
\label{tabs:table5}
\scalebox{0.75}{
\begin{tabular}{|c|c|c|c|c|c|c|c|c|}
\hline
\rowcolor[HTML]{D9E2F3}
\multicolumn{1}{|l|}{\multirow{2}{*}{}}                                                     & \multicolumn{4}{c|}{\textbf{InceptionV3}}                                   & \multicolumn{4}{c|}{\textbf{Dense-Net-121}}                                 \\ \cline{2-9} 
\rowcolor[HTML]{D9E2F3}
\multicolumn{1}{|l|}{}                                                                      & \textbf{Precision} & \textbf{Recall} & \textbf{F1-score} & \textbf{Support} & \textbf{Precision} & \textbf{Recall} & \textbf{F1-score} & \textbf{Support} \\ \hline
\textbf{\begin{tabular}[c]{@{}c@{}}Cercospora\_leaf\_spot \\ Gray\_leaf\_spot\end{tabular}} & 0.92               & 0.90            & 0.91              & 521              & 0.82               & 0.87            & 0.85              & 521              \\ \hline
\textbf{Common\_rust}                                                                       & 0.95               & 0.99            & 0.97              & 563              & 1.00               & 0.71            & 0.83              & 563              \\ \hline
\textbf{Northern\_Leaf\_Blight}                                                             & 0.94               & 0.94            & 0.94              & 561              & 0.96               & 0.64            & 0.77              & 561              \\ \hline
\textbf{Healthy}                                                                            & 1.00               & 0.99            & 0.99              & 550              & 0.63               & 1.00            & 0.77              & 550              \\ \hline
\textbf{Accuracy}                                                                           &                    &                 & 0.95              & 2195             &                    &                 & 0.80              & 2195             \\ \hline
\textbf{Macro avg}                                                                          & 0.95               & 0.95            & 0.95              & 2195             & 0.85               & 0.80            & 0.80              & 2195             \\ \hline
\textbf{Weighted avg}                                                                       & 0.95               & 0.95            & 0.95              & 2195             & 0.86               & 0.80            & 0.80              & 2195             \\ \hline

\rowcolor[HTML]{D9E2F3}
\multicolumn{1}{|l|}{\multirow{2}{*}{}}                                                     & \multicolumn{4}{c|}{\textbf{ResNet-101-V2}}                                 & \multicolumn{4}{c|}{\textbf{Xception}}                                      \\ \cline{2-9} 

\rowcolor[HTML]{D9E2F3}
\multicolumn{1}{|l|}{}                                                                      & \textbf{Precision} & \textbf{Recall} & \textbf{F1-score} & \textbf{Support} & \textbf{Precision} & \textbf{Recall} & \textbf{F1-score} & \textbf{Support} \\ \hline
\textbf{\begin{tabular}[c]{@{}c@{}}Cercospora\_leaf\_spot\\  Gray\_leaf\_spot\end{tabular}} & 0.98               & 0.32            & 0.48              & 521              & 0.97               & 0.74            & 0.84              & 521              \\ \hline
\textbf{Common\_rust}                                                                       & 0.80               & 0.97            & 0.88              & 563              & 0.88               & 0.98            & 0.93              & 563              \\ \hline
\textbf{Northern\_Leaf\_Blight}                                                             & 0.81               & 0.68            & 0.74              & 561              & 0.87               & 0.98            & 0.92              & 561              \\ \hline
\textbf{Healthy}                                                                            & 0.63               & 1.00            & 0.77              & 550              & 0.99               & 0.97            & 0.98              & 550              \\ \hline
\textbf{Accuracy}                                                                           &                    &                 & 0.75              & 2195             &                    &                 & 0.92              & 2195             \\ \hline
\textbf{Macro avg}                                                                          & 0.81               & 0.74            & 0.72              & 2195             & 0.93               & 0.92            & 0.92              & 2195             \\ \hline
\textbf{Weighted avg}                                                                       & 0.80               & 0.75            & 0.72              & 2195             & 0.93               & 0.92            & 0.92              & 2195             \\ \hline
\end{tabular}
}
\end{table}

\subsection{Experiment No 2 - Results - CNN Using Batch Normalization}
Over the years, there has been a significant improvement in terms of image-based plant detection approaches which have eased the process of image collection and have also made it cost-effective. However, despite such advancement, data annotation still remains a costly option which has created certain limitations for the availability of well-labeled datasets. In such situations, it becomes a difficult challenge for Machine learning models to classify the dataset accurately which impacts the accuracy of the overall predictions and classification. Thus, the current research work primarily aims to enhance the accuracy of plant disease classification by using a limited amount of training datasets. The dataset collected from these plants is not huge but it has a normal size. This dataset possesses multiple classes which are usually divided into two sets: testing set and training set. A total of 9279 images constitutes the image dataset used for disease recognition in corn and tomato leaves. We have used CNN with a layer of Batch Normalization for plant disaease identification.

In our study, we have employed CNN to serve as a framework for supporting our model for disease detection in corn and tomato. CNN is described as a subclass of deep artificial neural networks (ANNs) which is widely used around the globe for its efficiency and accuracy for tasks related to image classification of disease plants. CNN comprises various types of 2-D and 1-D layers in our model which include the following layers:

\begin{itemize}
    \item Convolutional layer   
    \item Pooling layers
    \item Batch normalization layers
    \item Dense layers
    \item Dropout layers
    \item Flatten layers
\end{itemize}

The convolutional layers are mainly involved in extracting certain attributes from the image dataset. The dimensionalities of these extracted images are reduced by the next layer, the pooling layer. The batch normalization layer is utilized to alleviate overfitting. It helps in the normalization of the convolutional layers and increases the pace of training and computation of ANNs. In our deep learning model, the fully connected layers are involved in collecting and compiling all data in previous layers to develop the final output.

For the appropriate functioning of CNN, the training dataset should be quite large. However, in most of the plant disease detection cases, the dataset is not usually large enough which creates problems for the accuracy of the prediction results. For instance, if the model parameters are greater in number than the number of samples, it would eventually lead to overfitting which would make the model an unreliable model for the accurate prediction of the disease. As our research experiment is on plant disease recognition, it is necessary to tackle this problem for the high accuracy of results. To this end, we have employed the data augmentation method.

\subsubsection{CNN Model Architecture} 
\textbf{Input Layer:} The input layer is employed to set up an instance of a Keras Tensor. A tensor is typically a generalized form of vectors and matrices to varying dimensions while Keras is referred to as a library of neural networks. Both frameworks, i.e., Keras and Tensors provide the opportunity to train the model with convenience. It is used as the first layer in this research project. The input layer returns (None, 128, 128, 3,) shape. In the input layer, we have processed the image having the resolution 128×128 px. The image is subjected to 3 filters. 

\textbf{Conv2d:} Conv2d layer comes after the input layer in our research project. This is basically a 2-dimensional layer that is mainly involved in creating a convolutional kernel. If one uses this layer as the first layer in the proposed model, the keyword ``input\_shape" should be utilized for the image in the available dataset. In case any pictures have variable sizes, use None. Three conv2d layers are used in this research. The first conv2d layer has the output dimension of (None, 64, 64, 64). It can be seen that resolution is reduced to 64 × 64 px and the number of filters used is increased to 64. The second conv2d layer returns the output shape of (None, 32, 32, 64), where image resolution is reduced to 32×32 px. However, the number of filters is kept the same (64). In the third conv2d layer, the resolution is further reduced to 8×8 and the number of channels is reduced to 32 returning an output shape of (None, 8, 8, 32). 

\textbf{Batch Normalization:} The fluctuation in the distribution of input layers is termed as internal covariate shift which creates certain challenges. In order to tackle this issue, batch normalization comes into play which is a widely used technique in DNNs to adjust inputs layers, assisting in a better learning process. In this research project, our model employs three batch normalization layers. It comes after the conv2d layer in the current model. Since it functions to standardize the inputs from the previous layers, all three batch normalization layers have the same output shape as that of previous conv2d layers. 

\textbf{Max Pooling:} The method is usually employed to select brighter patches from an image. Its importance becomes evident in case the image has a dark background and the interest of the researcher is the bright part of the image. In our research, after standardization and normalization of conv2d layers, these are sent to max pooling layers. These max pooling layers reduce the size of the image while keeping the number of filters unchanged. The first max pooling layer reduces the image resolution to 16×16 px returning the output shape (None, 16, 16, 64). The number of filters applied is 64. The second max pooling layer reduces the image resolution to 4×4 px while the number of filters is kept unchanged at 32 retuning the output shape (None, 4, 4, 32).  None simply means any batch size.

\textbf{Flatten:} Flatten layer simply converts the multidimensional image dataset into a 1-dimensional array. It then helps in inputting the data into the next layers to build to the neural network model. In our research project, flatten layer converts the multidimensional output from the previous max pooling layer (None, 4, 4, 32) into a single-dimensional output shape (None, 512).

\textbf{Dense:} It is an important neural network layer which refers to the fact that the dense layer is responsible for matrix-vector multiplication. Our model uses three dense layers after the flatten layer. This output dimension of the first dense layer is (None, 256) where 256 units are depicted. The second dense layer is used after the dropout layer with an output shape of (None, 128). The third dense layer depicts an output shape (None, 8) showing a reduction in layer units.

\textbf{Dropouts:} It is a technique that is mainly employed to alleviate the overfitting issues in ANN models. Dropout mainly functions by setting edges of the outgoing neurons. This layer has set the input units to 0. As a general practice, dropouts are employed after the dense layer of the neural network. Their use after the convolutional layers is usually avoided. In this research project, we have used 2 dropout layers after the dense layers. Its units remain the same as the outputs in the previous dense layer.

The architecture of the proposed CNN model for this research project consists of 15 layers having 7 different types of layers including one input layer, 3 convo 2D layers, 3 batch normalization layers, 2 max pooling layers, one flatten layer, 3 dense layers, and 2 dropout layers.

\subsubsection{Model Development}

\begin{figure*}[!ht]
\begin{center}
\includegraphics[width=0.50\textwidth]{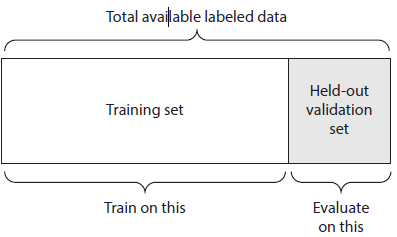}
\caption{Hold out cross-validation}
\label{figs:holdout}
\end{center}
\end{figure*}

\textbf{Hold out cross-validation method:} This method is an exhaustive cross-validation technique. We have used this technique to split the data into the training set and testing set along with a validation set. Thus, the data is split into three distinct parts. As a general rule, we have kept the splitting of the training set higher than the testing set by following an 80/20 splitting ratio. We have used this method to train the model by tuning its various parameters with the validation set. Then, during the final step, its performance is evaluated on the testing set (Figure~\ref{figs:holdout}).

\textbf{Hyper-parameters:} 
\begin{itemize} 
    \item Our model included 223,080 parameters among which 222,760 are found to be trainable.
    \item The image width is set at 128 while the image height is set at 128. 
\item Hence, the image size for the training model is 128×128.
\item The batch size is kept at 64. 
\item The number of classes for the training model is 8. 
\item The number of epochs is 50. 
\end{itemize}

\begin{figure*}[!ht]
\begin{center}
\includegraphics[width=0.6\textwidth]{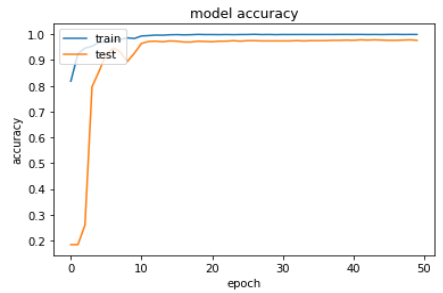}
\caption{Model Accuracy}
\label{figs:modelaccuracy}
\end{center}
\end{figure*}

\begin{figure*}[!ht]
\begin{center}
\includegraphics[width=0.6\textwidth]{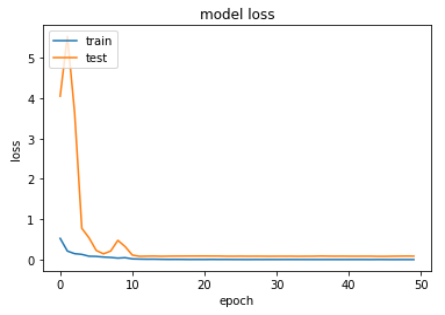}
\caption{Model Loss}
\label{figs:modelloss}
\end{center}
\end{figure*}

The results for the training model showed that all epochs showed high validation accuracy except epochs 1 and 2 with validation accuracy of 0.1853 and 0.2532, respectively which showed the highest validation loss. Epoch seven showed the highest accuracy (0.9375) followed by epoch 8 (8976). The model shows high overall validation accuracy for disease recognition with minimum validation loss. The graphical representation of the model accuracy is illustrated in Figure~\ref{figs:modelaccuracy}. The model loss is also depicted graphically in Figure~\ref{figs:modelloss}. Validation accuracies of epochs 35-49 are exhibited in the table which reveals the high prediction accuracy of our training model (Figure~\ref{figs:epochs} in Appendix).

 \subsubsection{Confusion Matrix:}
A confusion matrix is utilized to evaluate the accuracy of our disease categorization model.

\begin{itemize}
    \item The number of cases in which the prediction was a YES (Yes, the dataset showed the disease) and diseases was truly present refer to true positives. 
    \item Contrastingly, if the prediction is a NO, and the images truly do not show any diseases, this refers to true negatives. 
    \item If the during the classification, the prediction for the disease is a YES; however, no disease is detected in the image dataset, this refers to false positive or Type Two error. 
    \item Contrasting, if the prediction is a NO for the disease recognition; however, the disease is present, then this refers to false negatives or Type II error. 
\end{itemize}

\begin{figure*}[!h]
\begin{center}
\includegraphics[width=0.60\textwidth]{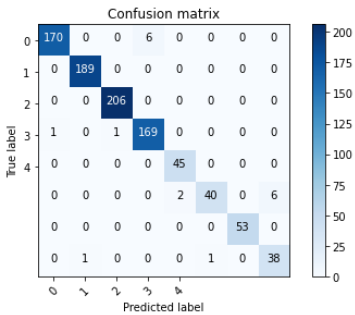}
\caption{Confusion matrix for the evaluation of classification performance}
\label{figs:confusion}
\end{center}
\end{figure*}

 The confusion matrix for our model is illustrated in Figure~\ref{figs:confusion}.

\subsection{Experiment 3} 

\subsubsection{Corn Leaves - Experiment}
The plantvillage dataset consists of images of leaves from various plants. The images of corn leaves are used for this experiment. There are 3 kinds of diseased leaves available in the dataset and they are individually classified along with healthy leaf images. So, the model classifies a given image in on of the 4 categories:

\begin{itemize}
    \item Healthy(1162 images)
\item Gray spot(513 images)
\item Common rust(1192 images)
\item Blight(985 images)
\end{itemize}

\textbf{Data Preparation:} The data-set is split into train, validation and test set in the ratio 0.6:0.2:0.2.

\begin{figure*}[!h]
\begin{center}
\includegraphics[width=0.95\textwidth]{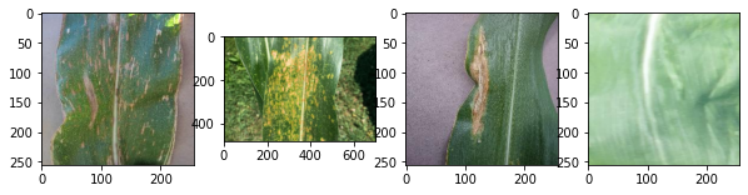}
\caption{Gray Spot, Common Rust, Blight, Healthy (Left to Right)}
\label{figs:corn123}
\end{center}
\end{figure*}

\textbf{Architecture Choice:} CNN architecture was used to classify the models.
\begin{itemize}
    \item CNN is the most commonly used architecture for image data.
\item CNNs can learn to extract color gradient details from the images. As the classification is mainly based on color, CNNs are a good choice.
\item The images are of size (256, 256, 3). So CNNs are most efficient as they can extract information and also reduce the size before passing data to ANN classifier for classification.
\item The exact architecture of each model is discussed in the next section. 
\end{itemize}

\textbf{Loss Function, Optimizer, Metrics:}

\begin{figure*}[!h]
\begin{center}
\includegraphics[width=0.35\textwidth]{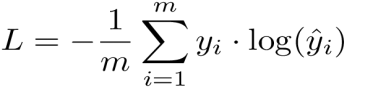}
\caption{Loss Function - Explanation}
\label{figs:987}
\end{center}
\end{figure*}

\begin{itemize}
    \item Loss Function: Cross entropy loss (Figure ~\ref{figs:987}). 

\item Optimizer: Adam, with a learning rate of 0.01 for all models.
\item Metrics: Accuracy, Confusion Matrix, Roc curve.
\end{itemize}

\textbf{Models}

\textbf{Model 1 - Baseline Model:} 
\begin{itemize}
   \item 6 convolution layers
\item 2 fully connected layers
\item Accuracy on test data: 95.5%
\item Confusion Matrix is shown in Figure~\ref{figs:cm1}.
\item ROC curves can be seen in Figure~\ref{fig:roc11},~\ref{fig:roc12},~\ref{fig:roc13},~\ref{fig:roc14}.
\item The model was able to classify healthy and common rust categories almost correctly. The results are expected because the model has seen these categories of images more compared to other two. The same trend can be observed in the following models too.

\end{itemize}

\textbf{Architecture:} Images for model 2 is similar to model 1 but each conv block is applied twice without resizing image from 256x256 to 64x64. For model 4, image is exactly similar to model 3(since l2 regularization is the only addon from model3).

\begin{figure*}[!h]
\begin{center}
\includegraphics[width=0.75\textwidth]{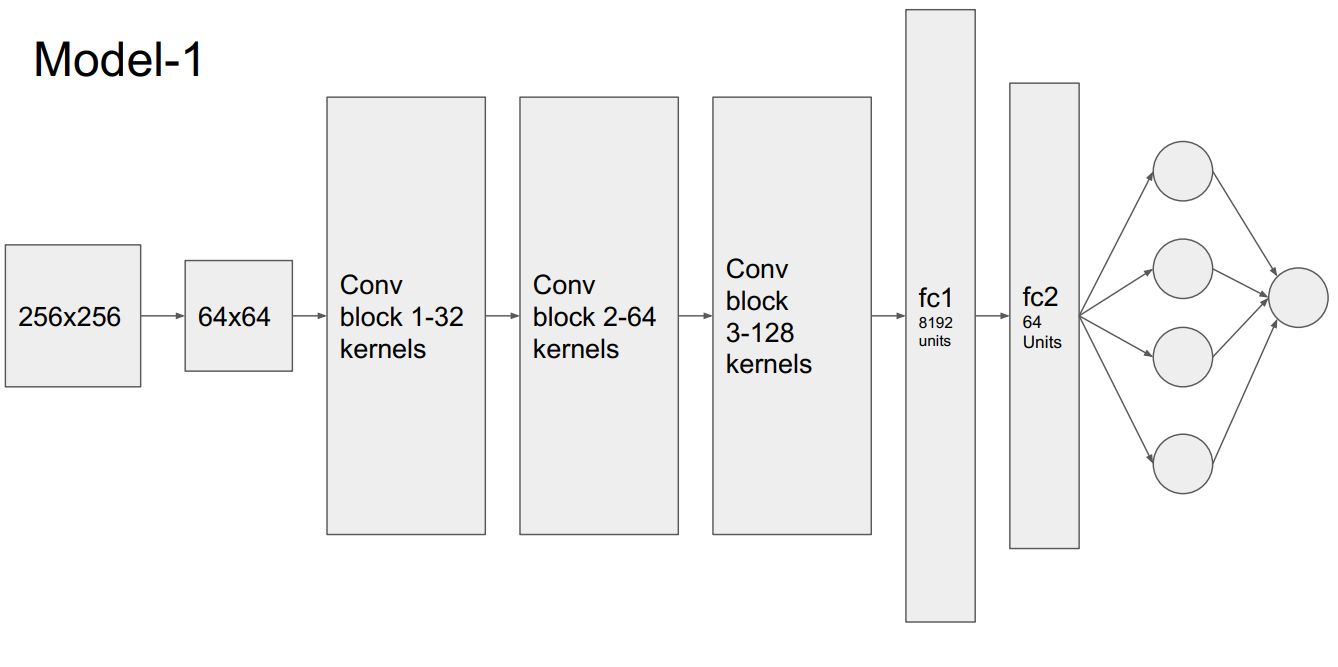}
\caption{Architecture Design Model 1}
\label{figs:model1}
\end{center}
\end{figure*}

\begin{figure*}[!h]
\begin{center}
\includegraphics[width=0.75\textwidth]{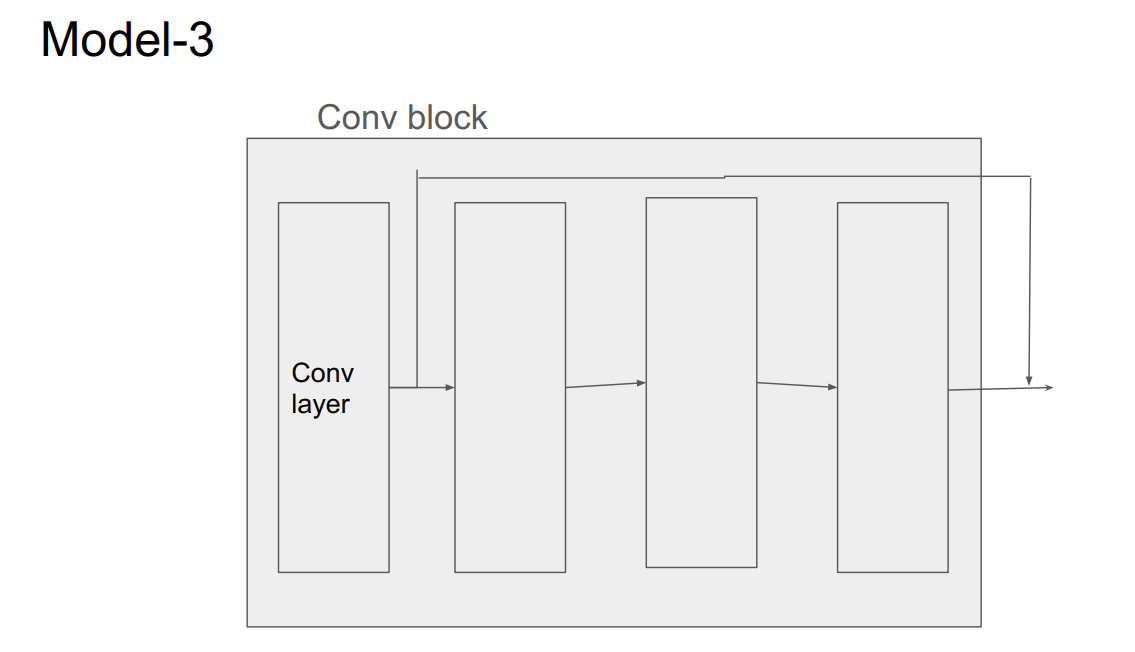}
\caption{Architecture Design for Model 3}
\label{figs:model3}
\end{center}
\end{figure*}

\begin{figure*}[h!]
\begin{center}
\includegraphics[width=0.71\textwidth]{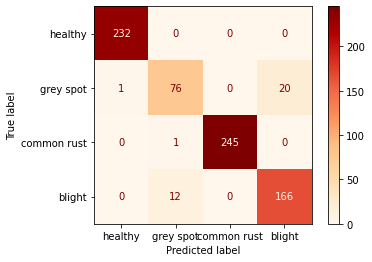}
\caption{Confusion Matrix}
\label{figs:cm1}
\end{center}
\end{figure*}

\begin{figure}
    \centering
    \begin{subfigure}[b]{0.47\textwidth}
        \includegraphics[width=\textwidth]{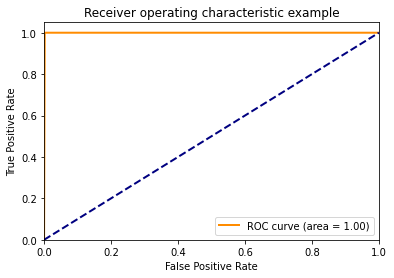}
\caption{Healthy Vs Not-Healthy}
\label{fig:roc11}
    \end{subfigure}
    ~ 
    \begin{subfigure}[b]{0.47\textwidth}
        \includegraphics[width=\textwidth]{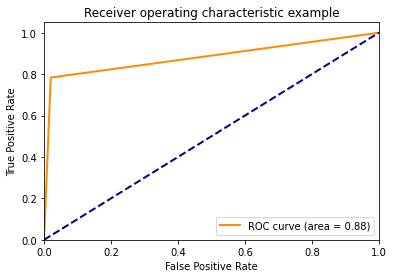}
\caption{Gray Spot Vs Not Gray Spot}
\label{fig:roc12}
    \end{subfigure}
         \caption{ROC Curves - 1}\label{fig:roc1}
\end{figure}

\begin{figure}
    \centering
    \begin{subfigure}[b]{0.47\textwidth}
        \includegraphics[width=\textwidth]{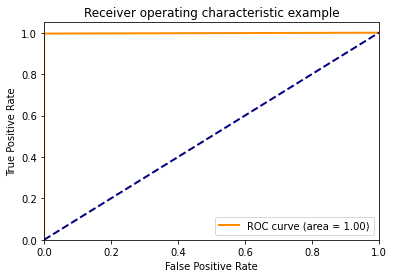}
\caption{Common Rust Vs Not Rust}
\label{fig:roc13}
    \end{subfigure}
    ~ 
    \begin{subfigure}[b]{0.47\textwidth}
        \includegraphics[width=\textwidth]{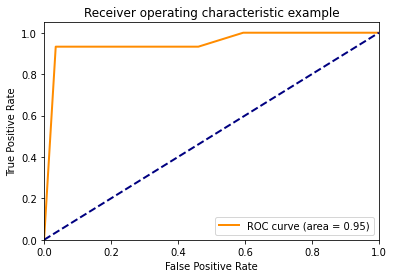}
\caption{Blight Vs Not Blight}
\label{fig:roc14}
    \end{subfigure}
         \caption{ROC Curves - 2}\label{fig:roc2}
\end{figure}

\begin{figure*}[!h]
\begin{center}
\includegraphics[width=0.55\textwidth]{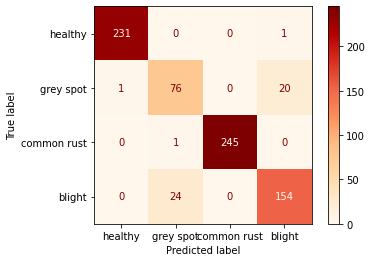}
\caption{Confusion Matrix}
\label{figs:cm2}
\end{center}
\end{figure*}

\begin{figure}
    \centering
    \begin{subfigure}[b]{0.47\textwidth}
        \includegraphics[width=\textwidth]{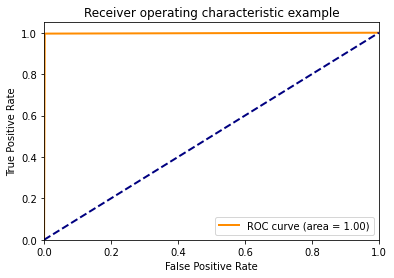}
\caption{Healthy Vs Not-Healthy}
\label{fig:roc21}
    \end{subfigure}
    ~ 
    \begin{subfigure}[b]{0.47\textwidth}
        \includegraphics[width=\textwidth]{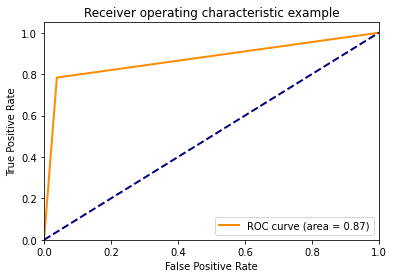}
\caption{Gray Spot Vs Not Gray Spot}
\label{fig:roc22}
    \end{subfigure}
         \caption{ROC Curves - 3}\label{fig:roc3}
\end{figure}

\begin{figure}
    \centering
    \begin{subfigure}[b]{0.47\textwidth}
        \includegraphics[width=\textwidth]{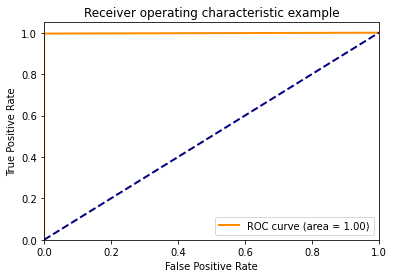}
\caption{Common Rust Vs Not}
\label{fig:roc23}
    \end{subfigure}
    ~ 
    \begin{subfigure}[b]{0.47\textwidth}
        \includegraphics[width=\textwidth]{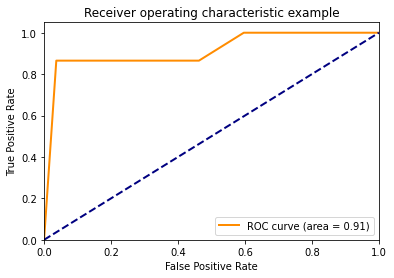}
\caption{Blight Vs Not }
\label{fig:roc24}
    \end{subfigure}
         \caption{ROC Curves - 4}\label{fig:roc4}
\end{figure}

\textbf{Model 2 - Increase of layers from baseline model:} 
\begin{itemize}
    \item 12 convolution layers
\item 4 fully connected layers
\item Accuracy on test data: 93.7\%
\item Confusion Matrix is shown in Figure~\ref{figs:cm2}.
\item ROC curves shown in Figure~\ref{fig:roc21},~\ref{fig:roc22},~\ref{fig:roc23},~\ref{fig:roc24}.
\item As we increase the model layers, the weights of initial layers will not be updated enough as the model trains. This happens because, when the model learns to identify some features, the gradient of loss function decreases. During back-propagation, as the gradient propagates, it will be multiplied with terms less than 1, further decreasing the value. Hence the initial layers remain the same. So the training loss does not decrease and the model doesn't perform well.
 
\end{itemize}

\begin{figure*}[!h]
\begin{center}
\includegraphics[width=0.55\textwidth]{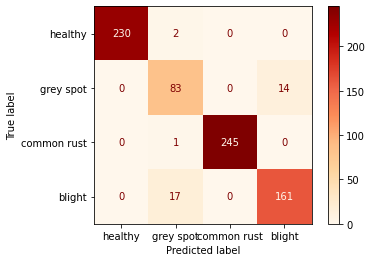}
\caption{Confusion Matrix}
\label{figs:cm3}
\end{center}
\end{figure*}

\begin{figure}
    \centering
    \begin{subfigure}[b]{0.47\textwidth}
        \includegraphics[width=\textwidth]{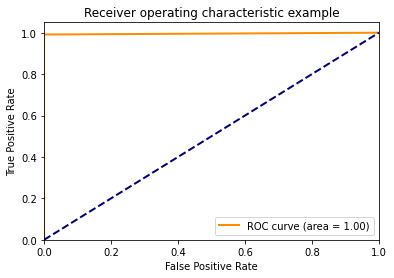}
\caption{Healthy Vs Not-Healthy}
\label{fig:roc31}
    \end{subfigure}
    ~ 
    \begin{subfigure}[b]{0.47\textwidth}
        \includegraphics[width=\textwidth]{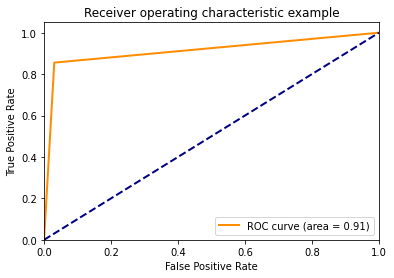}
\caption{Gray Spot Vs Not Gray Spot}
\label{fig:roc32}
    \end{subfigure}
         \caption{ROC Curves - 5}\label{fig:roc5}
\end{figure}

\begin{figure}
    \centering
    \begin{subfigure}[b]{0.47\textwidth}
        \includegraphics[width=\textwidth]{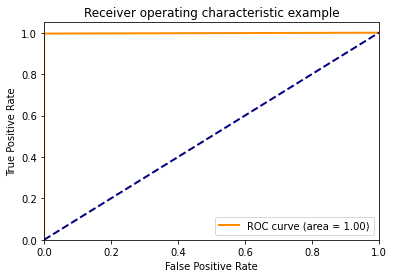}
\caption{Common Rust Vs Not}
\label{fig:roc33}
    \end{subfigure}
    ~ 
    \begin{subfigure}[b]{0.47\textwidth}
        \includegraphics[width=\textwidth]{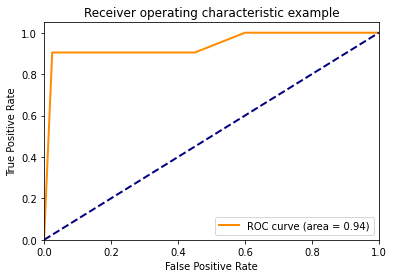}
\caption{Blight Vs Not Blight}
\label{fig:roc34}
    \end{subfigure}
         \caption{ROC Curves - 6}\label{fig:roc6}
\end{figure}

\textbf{Model 3 - Adding skip connections:} 
\begin{itemize}
    
\item 16 convolution layers
\item 2 fully connected layers
\item Skip/Residual connections for every 4 convolution layers
\item Accuracy on test data: 95.5%
\item Confusion Matrix is shown in the Figure~\ref{figs:cm3}.
\item ROC curves can be seen in the Figures~\ref{fig:roc5}, and ~\ref{fig:roc6}. 

\item The accuracy did not improve from the baseline model. But during training this model, we can observe that the training loss has gone down a lot more than validation loss, it in fact reached 0.01. This is an indication of over-fitting. The model remembers a little too much then generalizes the features from the training set. This happens because the dataset is imbalanced and the training dataset is relatively small wrt model complexity. This is another problem of complex models. They become a little more complex than required. But we cannot test every single hyperparameter and make the exact model we want, so we will introduce dropout layers in the model.

\end{itemize}

\begin{figure*}[!h]
\begin{center}
\includegraphics[width=0.55\textwidth]{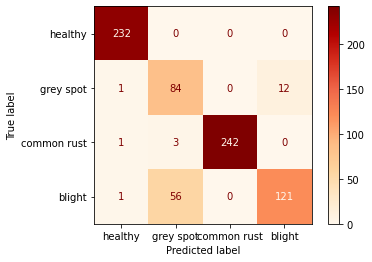}
\caption{Confusion Matrix}
\label{figs:cm4}
\end{center}
\end{figure*}

\begin{figure}
    \centering
    \begin{subfigure}[b]{0.47\textwidth}
        \includegraphics[width=\textwidth]{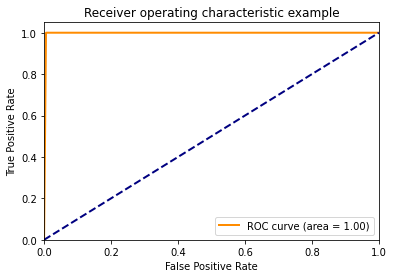}
\caption{Healthy Vs Not-Healthy}
\label{fig:roc41}
    \end{subfigure}
    ~ 
    \begin{subfigure}[b]{0.47\textwidth}
        \includegraphics[width=\textwidth]{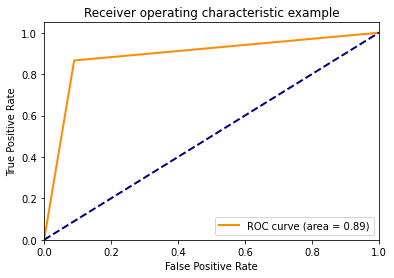}
\caption{Gray Spot Vs Not Gray Spot}
\label{fig:roc42}
    \end{subfigure}
         \caption{ROC Curves - 7}\label{fig:roc7}
\end{figure}

\textbf{Model 4 - Adding Regularization:}
\begin{itemize}
    \item When l2 regularizer is added to model3, the accuracy decreased to 91\%.
\item Confusion Matrix is shown in the Figure~\ref{figs:cm4}.

\item ROC curves can be seen in the Figure~\ref{fig:roc41}, Figure~\ref{fig:roc42}.

\item Adding a regularizer didn't help either. Some images of gray\_spot and blight are so similar to tell them apart, so more data of those kinds is required to model these features.

\end{itemize}

 \begin{figure}
    \centering
    \begin{subfigure}[b]{0.47\textwidth}
        \includegraphics[width=\textwidth]{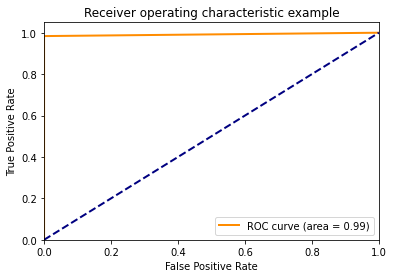}
\caption{Common Rust Vs Not}
\label{fig:roc43}
    \end{subfigure}
    ~ 
    \begin{subfigure}[b]{0.47\textwidth}
        \includegraphics[width=\textwidth]{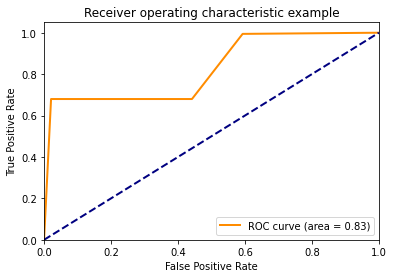}
\caption{Bright Vs Not}
\label{fig:roc44}
    \end{subfigure}
         \caption{ROC Curves - 8}\label{fig:roc8}
\end{figure}

\textbf{Summary:} When the hyper parameters of the model are optimized, it can be observed that they can classify healthy and common rust images with 100\% accuracy. Blight and Gray-spot diseases have many similarities,  so when a gray spot is smaller or the patches on the leaves affected due to blight are big, the model tends to misclassify the image. The combination of model1 with model3 would be the best model. As these NNs are based on probabilities(softmax function), images can be passed through both the models and the one with more confidence can be considered as a predicted label. When the models are trained with deeper models, the area under the roc curve is increased for Gray spot vs not roc curve. This can also be considered and use the confidence of a deeper model for the “Gray-spot” label.

 \begin{figure*}[!h]
\begin{center}
\includegraphics[width=0.71\textwidth]{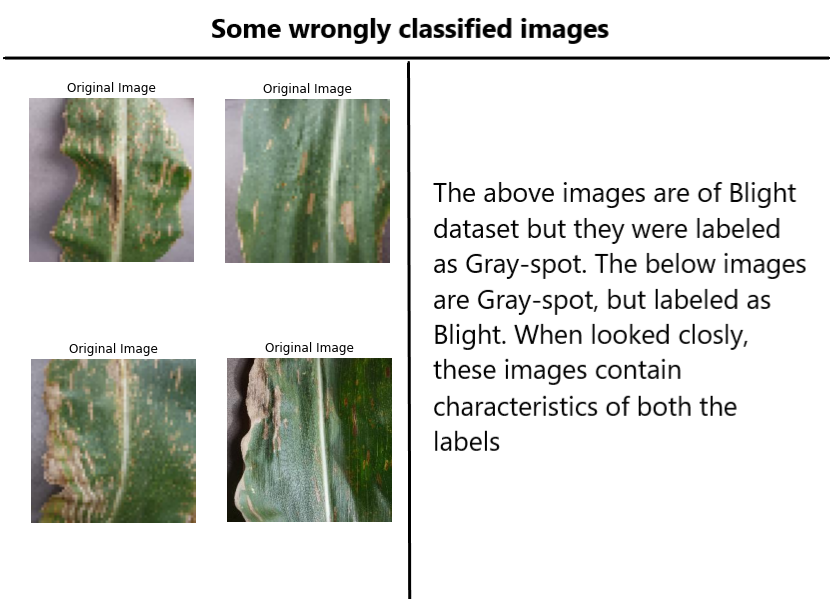}
\caption{Some Wrongly Classified Images}
\label{figs:summary}
\end{center}
\end{figure*}

\subsubsection{Tomatoes}

\begin{figure*}[!h]
\begin{center}
\includegraphics[width=0.99\textwidth]{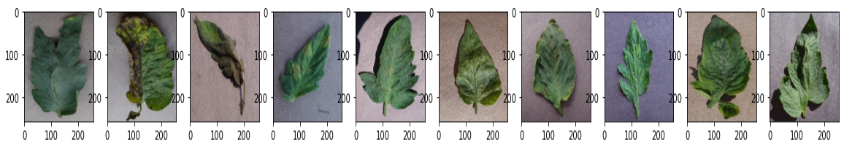}
\caption{Tomato Leaf Diseases Classification}
\label{figs:tldc}
\end{center}
\end{figure*}

\textbf{Dataset:} The plantvillage dataset consists of images of leaves from various plants. The images of tomato leaves are used for this experiment. There are 9 kinds of diseased leaves available in the dataset and they are individually classified along with healthy leaf images. So, the model classifies a given image in on of the 10 categories:

\begin{itemize}
    \item Healthy
\item Bacterial spot
\item Early blight
\item Late blight
\item Leaf mold
\item Septoria leaf spot
\item Spider mites
\item Target spot
\item Mosaic virus
\item Yellow Leaf curl virus
\item The image of these diseases can be seen in the Figure~\ref{figs:tldc}.
\end{itemize}

\textbf{Data Preparation:} 
The dataset is split into train, validation and test set in the ratio 0.7:0.2:0.1.

\textbf{Architecture Choice:} CNN architecture was used to classify the models. 

\begin{itemize}
    \item CNN is the most commonly used architecture for image data.
\item CNNs can learn to extract color gradient details from the images. As the classification is mainly based on color, CNNs are a good choice.
\item The images are of size (256, 256, 3). So CNNs are most efficient as they can extract information and also reduce the size before passing data to ANN classifier for classification.
\item The exact architecture of each model is shown below.
\end{itemize}

\textbf{Loss Function, Optimizer, Metrics:}

\begin{itemize}
    \item Loss Function: Cross entropy loss (~\ref{figs:987}. 

\item Optimizer: Adam, with a learning rate of 0.01 for all models.
\item Metrics: Accuracy, Confusion Matrix, Roc curve.
\end{itemize}

\textbf{Models}

\begin{figure*}[!h]
\begin{center}
\includegraphics[width=0.85\textwidth]{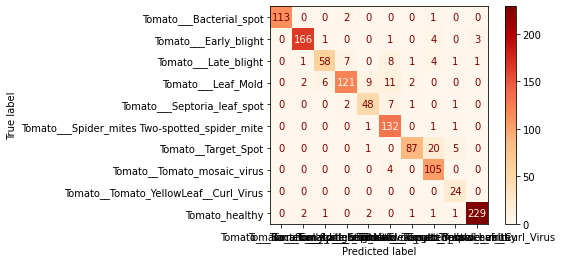}
\caption{Confusion Matrix}
\label{figs:tcm1}
\end{center}
\end{figure*}

\textbf{Model 1 - Baseline model:}
\begin{itemize}
    \item 6 convolution layers
\item 2 fully connected layers
\item Accuracy on test data: 90.25\%.
\item Confusion Matrix is shown in the Figure~\ref{figs:tcm1}.
\item Model 1 is the baseline model with less layers. This model has a lot of confusion among classes. A model with more no. of layers might fit the data better.
\end{itemize}

\textbf{Model 2 - Increase of layers from baseline model:} 

\begin{figure*}[!h]
\begin{center}
\includegraphics[width=0.85\textwidth]{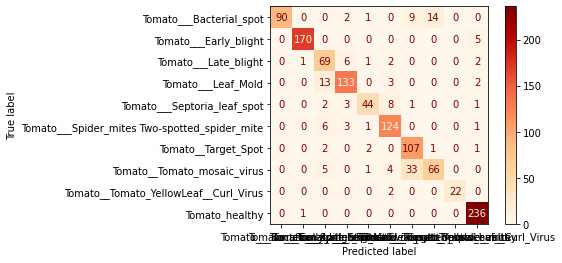}
\caption{Confusion Matrix}
\label{figs:tcm2}
\end{center}
\end{figure*}

\begin{itemize}
    \item 9 convolution layers
\item 4 fully connected layers
\item Accuracy on test data: 88.87%
\item Confusion Matrix can be seen in the Figure~\ref{figs:tcm2}.
\item As we increased the model complexity, the weight gradients for initial layers would become zero. As a result, the model is performing poorly. This can be fixed by using skip connections.

\end{itemize}

\textbf{Model 3 - Adding skip connections:} 

\begin{figure*}[!h]
\begin{center}
\includegraphics[width=0.85\textwidth]{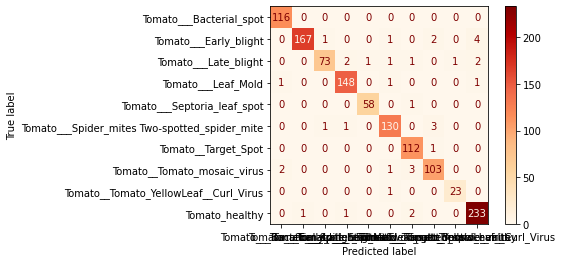}
\caption{Confusion Matrix}
\label{figs:tcm3}
\end{center}
\end{figure*}

\begin{itemize}
    \item 12 convolution layers
\item 2 fully connected layers
\item Skip/Residual connections for every 4 convolution layers
\item Accuracy on test data: 97%
\item Confusion Matrix can be seen in the Figure~\ref{figs:tcm3}.
\item Adding skip connections is making the model perform very well. Only those images which are very less diseased were classified wrongly by this model. We can see that the model performs very well compared to the similar model used for corn dataset. This is because of the increased amount of images per label(for some labels) for the tomato dataset. 
\end{itemize}

\textbf{Summary:} The tomato dataset contains more data w.r.t label for 4 labels when compared to corn dataset. As a result, when we have an optimized model the accuracy is more than the accuracy for corn data. We can also observe that the model with more layers is performing well for the tomato dataset whereas it was not performing well(on test data) for corn data. This is an indication of overfitting due to lack of availability of data.

\subsection{Experiment 4}

In this experiment, all the images of Corn and tomatos were combined and further run different model. Dataset 17603 were used for training 4409 images for testing purposes. Furthermore, different models such as, MobileNet, EfficientNetB0, Xception, InceptionResNetV2, and CNN  accuracy and loss graphs are discussed. Besides that, we have also proposed our model and its efficiency and loss, the detail of the model can be read below.

\subsubsection{MobileNet Model}

\begin{figure}
    \centering
    \begin{subfigure}[b]{0.47\textwidth}
        \includegraphics[width=\textwidth]{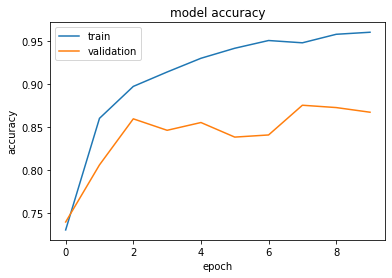}
\caption{MobileNet Model  Accuracy Graph}
\label{fig:3e_MN_ACC}
    \end{subfigure}
    ~ 
    \begin{subfigure}[b]{0.47\textwidth}
        \includegraphics[width=\textwidth]{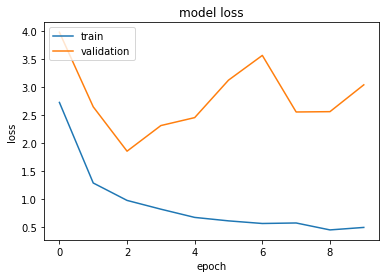}
\caption{MobileNet Model  Loss Graph}
\label{fig:3e_MN_LOSS}
    \end{subfigure}
         \caption{MobileNet Model}\label{fig:mobile}
\end{figure}

MobileNet, which employs depth-wise separable convolutions compared to a network with conventional convolutions of the same depth in the nets \footnote{it dramatically reduces the number of parameters}. Lightweight deep neural networks are the outcome of this. The Mobile Net model is loaded, the necessary parameters are configured, and the model is trained using the training dataset over ten epochs. It is possible to deduce from the results that the training data had an accuracy score of 96\% and the test dataset had an accuracy score of 87\%. The training and testing accuracy and loss plots are shown in the Figure~\ref{fig:mobile}.

\subsubsection{EfficientNetB0 Model}

\begin{figure}
    \centering
    \begin{subfigure}[b]{0.47\textwidth}
        \includegraphics[width=\textwidth]{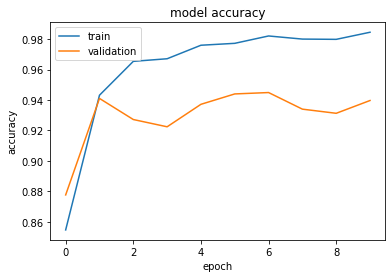}
\caption{EfficientNetB0 Model  Accuracy Graph}
\label{fig:3e_EN_ACC}
    \end{subfigure}
    ~ 
    \begin{subfigure}[b]{0.47\textwidth}
        \includegraphics[width=\textwidth]{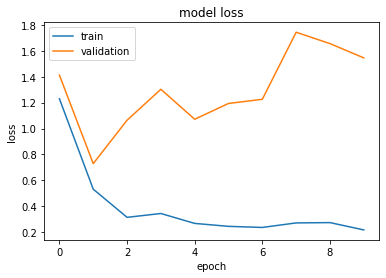}
\caption{EfficientNetB0 Model  Loss Graph}
\label{fig:3e_EN_LOSS}
    \end{subfigure}
         \caption{EfficientNetB0 Model}\label{fig:effinet}
\end{figure}

The EfficientNetB0 model (a convolutional neural network architecture) is applied in the next step, which employs a compound coefficient to scale all depth, width, and resolution dimensions evenly. The implementation in this paper uses the EfficientNetB0 pre-trained model. The required settings are set after the pre-trained model has been initialized, including the weights set to the image net and the top layer set to False, among other necessary parameters. The EfficientNetB0 model is then trained using the training dataset, and its performance is assessed using the test dataset. From the results, it can be deduced that the training dataset's accuracy score is 98.44 percent, whereas the test dataset's accuracy score is 95 percent. The training and testing accuracy and loss plots are shown in the Figure~\ref{fig:effinet}.

\subsubsection{Xception Model}

\begin{figure}
    \centering
    \begin{subfigure}[b]{0.47\textwidth}
        \includegraphics[width=\textwidth]{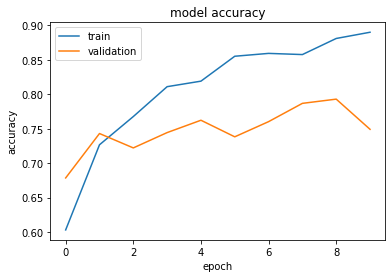}
\caption{Xception Model  Accuracy Graph}
\label{fig:3e_XC_ACC}
    \end{subfigure}
    ~ 
    \begin{subfigure}[b]{0.47\textwidth}
        \includegraphics[width=\textwidth]{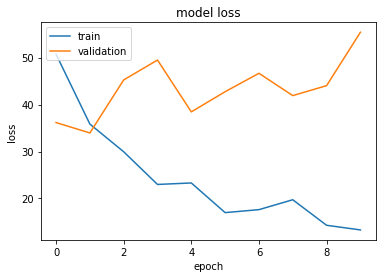}
\caption{Xception Model  Loss Graph}
\label{fig:3e_XC_LOSS}
    \end{subfigure}
         \caption{Xception Model}\label{fig:xcept}
\end{figure}

With Depthwise Separable Convolutions, the deep convolutional neural network architecture known as Xception is used ( which has 71 layers in total). This study uses an Xception model that has already been trained using more than a million photos from the ImageNet database. The model is developed using the training dataset. In this study, data is used for training in the proportion of 80\% and testing in the proportion of 20\%. It is possible to deduce from the results that the test dataset had an accuracy score of 75\%, and the training data had an accuracy score of 89\%. The training and testing accuracy and loss plots are shown in the Figure~\ref{fig:xcept}.

\subsubsection{InceptionResNetV2 Model}
\begin{figure}
    \centering
    \begin{subfigure}[b]{0.47\textwidth}
        \includegraphics[width=\textwidth]{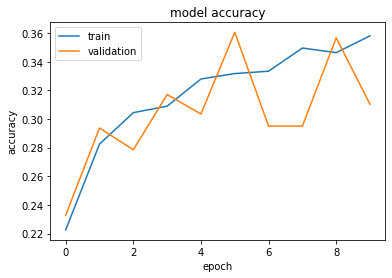}
\caption{InceptionResNetV2 Model  Accuracy Graph}
\label{fig:3e_IRV2_ACC}
    \end{subfigure}
    ~ 
    \begin{subfigure}[b]{0.47\textwidth}
        \includegraphics[width=\textwidth]{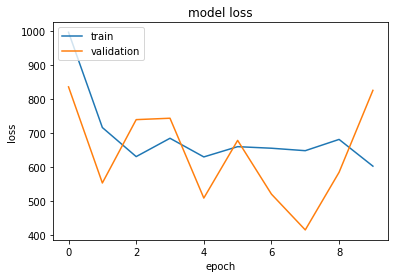}
\caption{InceptionResNetV2 Model  Loss Graph}
\label{fig:3e_IRV2_LOSS}
    \end{subfigure}
         \caption{InceptionResNetV2 Model}\label{fig:incep}
\end{figure}

A convolutional neural architecture called Inception-ResNet-v2 expands on the Inception family of architectures while incorporating residual connections (replacing the filter concatenation stage of the Inception architecture). The network has 164 layers and was trained using the imagenet database's millions of photos. In this study, a pre-trained InceptionResNetV2 model is taken into consideration for implementation. Once the InceptionResNetV2 pre-trained model has been initialized and the weights have been set to imagenet. The training dataset is used to create the model, and the test dataset is used to assess the model's performance. The results can be deduced that the training dataset yielded an accuracy score of 35.81 percent, whereas the test dataset yielded a score of 31.03 percent. The training and testing accuracy and loss plots are shown in the Figure~\ref{fig:incep}.

\subsubsection{Convolutional Neural Network (CNN)}

\begin{figure}
    \centering
    \begin{subfigure}[b]{0.47\textwidth}
        \includegraphics[width=\textwidth]{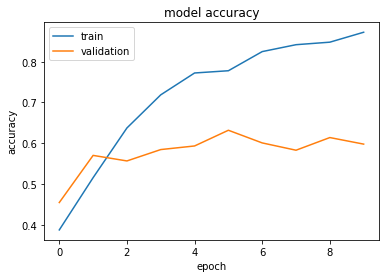}
\caption{Convolutional Neural Network Model Accuracy Graph}
\label{fig:3e_CNN_ACC}
    \end{subfigure}
    ~ 
    \begin{subfigure}[b]{0.47\textwidth}
        \includegraphics[width=\textwidth]{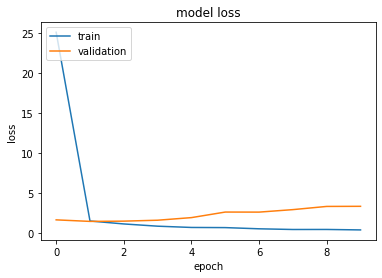}
\caption{Convolutional Neural Network Model Loss Graph}
\label{fig:3e_CNN_LOSS}
    \end{subfigure}
         \caption{Convolutional Neural Network Model}\label{fig:cnnm}
\end{figure}

One of the main deep learning architectures is Convolutional Neural Networks (CNN). In this research, along with other deep learning models, Convolutional Neural Network is also implemented. As the model is implemented layer by layer. So in the first step, the model is defined as sequential, after this there are  two Conlution 2 D layers, maxpooling layer and in the end the Dense Neural Network is implemented. After compiling the model. The model is training on the training dataset and the performance of the model is evaluated on the test dataset. From the results it can be analyzed that an accuracy score of 87.24 percent is obtained on the training data and an accuracy 60 percent is obtained on the test dataset. The training and testing accuracy and loss plots are shown in the Figure~\ref{fig:cnnm}.

\subsubsection{Our Proposed Model}

\begin{figure*}[!h]
\begin{center}
\includegraphics[width=0.55\textwidth]{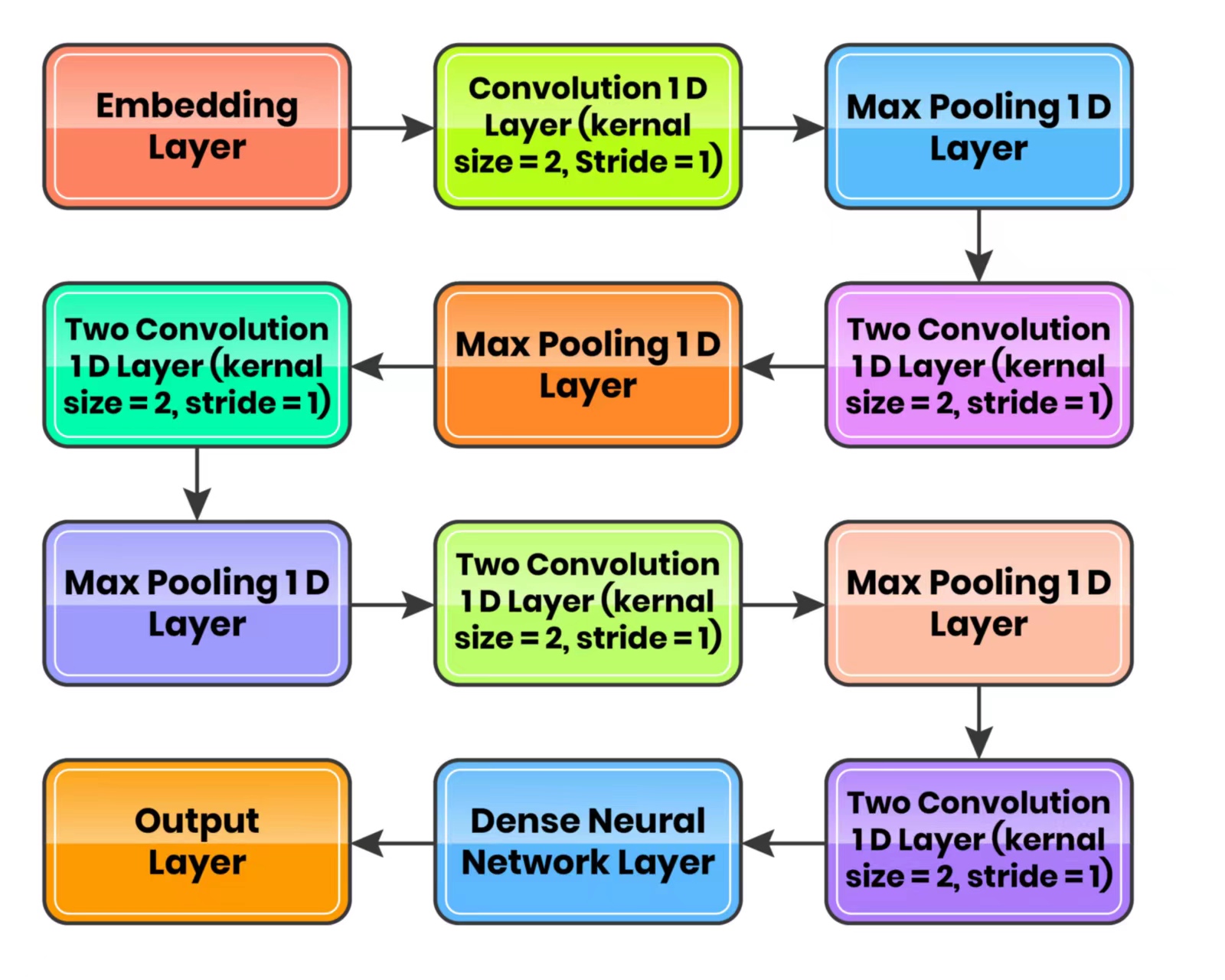}
\caption{Proposed Model Complete Architecture}
\label{figs:marchi}
\end{center}
\end{figure*}

\begin{figure}
    \centering
    \begin{subfigure}[b]{0.47\textwidth}
        \includegraphics[width=\textwidth]{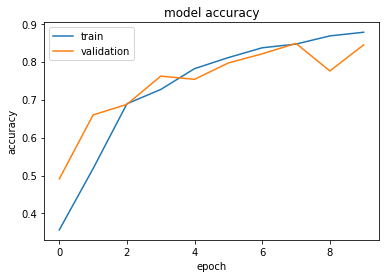}
\caption{Proposed Model Accuracy Graph}
\label{fig:3e_PA_ACC}
    \end{subfigure}
    ~ 
    \begin{subfigure}[b]{0.47\textwidth}
        \includegraphics[width=\textwidth]{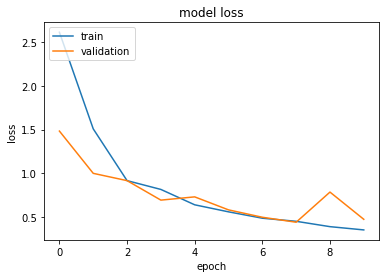}
\caption{Proposed Model Loss Graph}
\label{fig:3e_PA_LOSS}
    \end{subfigure}
         \caption{Proposed Model}\label{fig:pmm}
\end{figure}

Based on the convolutional neural network, a novel approach is proposed in this paper, the most unique thing about the proposed approach is that it focuses on having convolution layers of 6 x 6 filter along with a stride of 1  and always the same padding as well as max pooling layer of 2x2 filter of stride 3 is used. The process is followed through the complete architecture (Figure~\ref{figs:marchi}) i.e. in the complete architecture, all the convolution layers of 6x6 filter along with a stride of 1 are implemented and the same padding and max pooling layer of 2x2 filter with a stride of 3 is used in the complete architecture. After initializing the model, the model is trained on the training dataset and the performance of the model is evaluated on the test dataset. 80 percent of the data is used for the training purpose and 20 percent of the data is used for the testing purpose. From the results it can be analyzed that an accuracy score of  87.72 percent is obtained on the training dataset and an accuracy of 84.42 percent is obtained on the test dataset. This shows that our proposed model depicts a very satisfactory performance on the training dataset and on the test dataset and the issues of overfitting and underfitting are not faced. The training and testing accuracy and loss plots are shown in the Figure~\ref{fig:pmm}. The comparison of the models discussed above can be seen in the Table~\ref{tabs:comp}.

In Experiment 4, we propose a novel strategy based on convolutional neural networks. The approach's most distinctive feature is that it emphasizes using convolution layers of a 6 x 6 filter along with a stride of 1 and the same padding as well as a max pooling layer of a 2x2 filter of stride 3. The procedure is carried out across the entire architecture (Figure~\ref{figs:marchi}), i.e., all of the 6x6 convolution layers with a stride of 1 are implemented, and the same padding and max pooling layer of 2x2 filter with a stride of 3 is utilized. After initialization, the model is trained using the training dataset, and its effectiveness is assessed using the test dataset. Data is used for training purposes in an 80/20 ratio, with 20\% of the data being used for testing. From the results, it can be deduced that the training dataset yielded an accuracy score of 87.72 percent, whereas the test dataset yielded an accuracy score of 84.42 percent. This demonstrates that the concerns of overfitting and underfitting are not present, and our proposed model exhibits excellent performance on both the training and test datasets. The Figure~\ref{fig:pmm} displays the accuracy and loss graphs for training and testing. The Table~\ref{tabs:comp} shows a comparison of the models outlined previously.

\begin{table}[h!]
\centering
\caption{Comparison of Model Efficiency}
\label{tabs:comp}
\scalebox{0.75}{
\begin{tabular}{|l|c|c|c|c|c|}
\hline
\rowcolor[HTML]{96FFFB} 
\multicolumn{1}{|c|}{\cellcolor[HTML]{96FFFB}\textbf{Model}} & \textbf{Dataset}                                                 & \textbf{Accuracy} & \textbf{Val\_Accuracy} & \textbf{Loss} & \textbf{Val\_Loss} \\ \hline
\textbf{CNN}                                                 & \begin{tabular}[c]{@{}c@{}}Plant Village \\ Dataset\end{tabular} & 87.24\%           & 59.79\%                & 0.4336        & 3.3793             \\ \hline
\textbf{InceptionReSNetV2}                                   & \begin{tabular}[c]{@{}c@{}}Plant Village \\ Dataset\end{tabular} & 35.81\%           & 31.03\%                & 602           & 825.44             \\ \hline
\textbf{Mobile Net}                                          & \begin{tabular}[c]{@{}c@{}}Plant Village \\ Dataset\end{tabular} & 96.07\%           & 86.73\%                & 0.4889        & 3.0399             \\ \hline
\textbf{Efficient Net}                                       & \begin{tabular}[c]{@{}c@{}}Plant Village \\ Dataset\end{tabular} & 98.44\%           & 93.97\%                & 0.2170        & 1.5455             \\ \hline
\textbf{Xception}                                            & \begin{tabular}[c]{@{}c@{}}Plant Village \\ Dataset\end{tabular} & 89\%              & 74.91\%                & 13.3072       & 55.4075            \\ \hline \hline
\textbf{\textit{Proposed Approach}}                                   & \begin{tabular}[c]{@{}c@{}}Plant Village \\ Dataset\end{tabular} & 87.75\%           & 84.42\%                & 0.3498        & 0.4713             \\ \hline \hline
\end{tabular}
}
\end{table}

\section{Conclusion}
There are numerous automated or computer-oriented methods for detecting and categorizing plant diseases; however, this research field is undeveloped. In addition, no commercial solutions are available save those dealing with plant-based identification of plant species.

This article aims to examine a new method to use deep learning to identify and diagnose plants’ illnesses using plant images automatically. The developed model identified leaves and differentiated between two plant types (tomato, maize) with healthy leaves and six unique visually identifiable diseases.It was explained in detail, beginning with gathering pictures for training and validation, then pre-processing and expanding the image, and finally, the CNN. The newly proposed model was then put through trials to check its effectiveness.

We have used photographs from the PlantVillage dataset to construct a new plant disease image collection with images of 1963 tomato plants. 1374 photos were utilized for training, while 589 images were used for testing and validation. Also, we have 7316 photos of corn plants extracted from the PlantVillage dataset, of which 5121 were utilized for training, and 2195 were used for testing/validation. The experimental results achieved we have evaluated that in the four models that we have used, the Xception model has better val\_accuracy and val\_loss than all other models for the tomato and corn dataset. The evaluated values for the tomato and corn dataset are ``val\_accuracy = 95.08\%, and val\_los = 0.3108", and ``val\_accuracy = 92.21\%, and val\_los = 0.4204" respectively. In another experiment, we have utilized CNN using Batch Normalization for disease detection in Tomatoes and Corn leaves. The results for detection of diseases for both the vegetable leaves are around 99.90\% (training set) and for val\_accuracy= 98.0\% with val\_loss= 0.103. 
In experiment number 3, CNN architecture was used to classify as the base model. As a next step, more layers were added in the baseline model (Model 2). Furthermore, skip connections were added in model 3, and in model 4, regularizations were added. The detail of the experiment and models efficicency is shown in the paper (sub-section 1.5).  In experiment number 4, all the images of corn and tomato's were combined and further run different model. Dataset of 17603 were used for training, and 4409 images for testing purposes. Furthermore, different models such as, MobileNet (val\_accuracy= 86.73\%), EfficientNetB0 (val\_accuracy= 93.973\%), Xception (val\_accuracy= 74.91\%), InceptionResNetV2 (val\_accuracy= 31.03\%), and CNN (59.79\%)  were used. Besides that, we have also proposed our model with val\_accuracy= 84.42\%.

An expansion of this research is to collect images to expand the dataset and enhance the model accuracy via various methods.

\textbf{Future Work:} Future research aims to create a comprehensive system of server-side components that includes an advanced prototype and an application for intelligent portable devices with features identifying leaf diseases of fruit, vegetables, and other plants taken with a cell phone camera. The solution will help farmers identify plant diseases quickly and effectively and facilitate decision-making processes for chemical pesticides (independent of the degree of expertise). In addition, future studies will include the expansion of the model by training it to recognize plants in larger regions, integrating aerial images of oak and wine-growing fields taken by drones with neural networks for objects identification. 

\vspace{3mm}



 
  
 \vspace{3mm}

  \textbf{Data Availability Statement:} Dataset for the research paper can be found at the link \url{https://www.kaggle.com/abdallahalidev/plantvillage-dataset}.


\bibliographystyle{elsarticle-num}

\bibliography{elsevier}

\end{document}